\documentclass[lettersize,journal]{IEEEtran}

\usepackage[utf8]{inputenc}
\usepackage[T1]{fontenc}
\usepackage{csquotes}
\usepackage{color}
\usepackage[pdftex,dvipsnames]{xcolor}
\usepackage[]{graphicx}
\usepackage{hyperref}
\usepackage{xargs}
\usepackage{wrapfig}
\usepackage{textcomp}
\usepackage{bm}
\usepackage{amsmath}
\usepackage{gensymb}
\usepackage{booktabs}
\usepackage{adjustbox}
\usepackage{float}
\usepackage{helvet}
\usepackage[caption=false,font=normalsize,labelfont=sf,textfont=sf]{subfig}
\newcommand*{\figtiTFlitefont}{\fontfamily{phv}\selectfont}

\usepackage{xfrac}

\usepackage[ruled,vlined]{algorithm2e}

\newcommand{\etal}{\textit{et al.}}
\newcommand{\cmt}[1]{\ignorespaces}

\newcommand{\halfwidth}{8.5cm}
\newcommand{\fullpath}[1]{./img/#1}

\usepackage{tikz}

\newcommand*\annotatedFigureText[4]{\node[draw=none, anchor=south west, text=#2, inner sep=0, text width=#3\linewidth,font=\sffamily] at (#1){#4};}
\newenvironment {annotatedFigure}[1]{\centering\begin{tikzpicture}
    \node[anchor=south west,inner sep=0] (image) at (0,0) { #1};\begin{scope}[x={(image.south east)},y={(image.north west)}]}{\end{scope}\end{tikzpicture}}

\newcommand{\ratio}{S_{P100}}

\newcommand{\meff}{\eta_{m}}

\newcommand{\tf}{T_{i}}
\newcommand{\power}{P_w}
\newcommand{\energy}{E_i}

\newcommand{\art}{state-of-the-art}

\usepackage{multirow}
\usepackage{bigstrut}
\usepackage{makecell}

\newcommand{\bferra}{Bruno Ferrarini}

\newcommand{\bferraEm}{bferra}

\newcommand{\sheh}{Shoaib Ehsan}

\newcommand{\shehEm}{sehsan}

\newcommand{\kdm}{Klaus D. McDonald-Maier}

\newcommand{\kdmEm}{kdm}

\newcommand{\mm}{Michael Milford}
\newcommand{\mmEM}{michael.milford@qut.edu.au}

\begin{document}

\title{Binary Neural Networks for Memory-Efficient and Effective Visual Place Recognition in Changing Environments}

\author{\bferra{}, \mm{}, \kdm{} and \sheh{} 
\thanks{Manuscript received June 20, 2021; revised October 1, 2021; accepted January
20, 2022. Date of publication \textbf{PLACEHOLDER}; date of current version December \textbf{PLACEHOLDER}. This work was supported by the UK Engineering and Physical Sciences
Research Council through grants EP/R02572X/1, EP/P017487/1, and in part by the RICE project funded by the National Centre for Nuclear Robotics Flexible Partnership Fund. This paper was recommended for publication by Associate Editor
\textbf{PLACEHOLDER} and Editor Francois Chaumette upon evaluation of the reviewers’ comments. \textit{(Corresponding author: \bferra{}.)}} 
\thanks{\bferra{}, \kdm{} and \sheh{} are with the School of Computer Science and Electronic Engineering, University of Essex, Colchester, CO4 3SQ, UK
        {\tt\small \{\bferraEm{},\kdmEm{}, \shehEm{}\}@essex.ac.uk}}%
\thanks{\mm{} is with the QUT Centre for Robotics, School of Electrical Engineering and Robotics and the Australian Centre for Robotic Vision at the Queensland University of Technology, and was partially supported by ARC grants FT140101229, CE140100016 and the QUT Centre for Robotics.
        {\tt\small \mmEM{}}}%
\thanks{Digital Object Identifier: \textbf{PLACEHOLDER}}
}

\markboth{IEEE Transaction on Robotics. Preprint Version. Accepted January, 2022}%
{Ferrarini \etal{}: Binary Neural Networks for Memory-Efficient and Effective VPR.}
\IEEEpubid{\makebox[\columnwidth]{\copyright~2022 IEEE. Personal use of this material is permitted\hfill} \hspace{\columnsep}\makebox[\columnwidth]{ }}

\maketitle

\begin{abstract}

Visual place recognition (VPR) is a robot's ability to determine whether a place was visited before using visual data. While conventional hand-crafted methods for VPR fail under extreme environmental appearance changes, those based on convolutional neural networks (CNNs) achieve state-of-the-art performance but result in heavy runtime processes and model sizes
that demand a large amount of memory. Hence, CNN-based approaches are unsuitable for resource-constrained platforms, such as small robots and drones. In this paper, we take a multi-step approach of decreasing the precision of model parameters, combining it with network depth reduction and fewer neurons in the classifier stage to propose a new class of highly compact models that drastically reduces the memory requirements and computational effort while maintaining state-of-the-art VPR performance. To the best of our knowledge, this is the first attempt to propose binary neural networks for solving the visual place recognition problem effectively under changing conditions and with significantly reduced resource requirements. Our best-performing binary neural network, 
dubbed FloppyNet, achieves comparable VPR performance when considered against its full-precision and deeper counterparts while consuming 99\% less memory and increasing the inference speed seven times.

\end{abstract}


\section{Introduction}

\label{sec:intro}
\IEEEPARstart{V}{isual} place recognition addresses the problem of determining whether a location has been visited before using visual information. VPR is a fundamental task for autonomous navigation. It enables a robot to re-localize itself in the workspace when the position tracking fails or drifts due to accumulated errors. However, variations in viewpoint and appearance due to seasonal, weather and illumination changes render VPR challenging for mobile robots. While conventional hand-crafted techniques for VPR fail under extreme environmental changes, those based on deep CNNs achieve \art{} performance \cite{zaffar2020vpr} but result in heavy runtime processes and model sizes that demand a large amount of memory.

Mobile robots are often equipped with resource-constrained hardware that limits the usability of such demanding techniques
\cite{maffra2018tolerant,ferrarini2019visual}. Increasing the efficiency by saving memory and reducing the computational effort to run a model without sacrificing the performance is paramount for such resource-constrained mobile robots. Higher efficiency enables VPR on cheap hardware and frees resources for additional functionalities to improve a robot's navigation system.

\begin{figure}[pt]
\centering
\includegraphics[width=7.8cm]{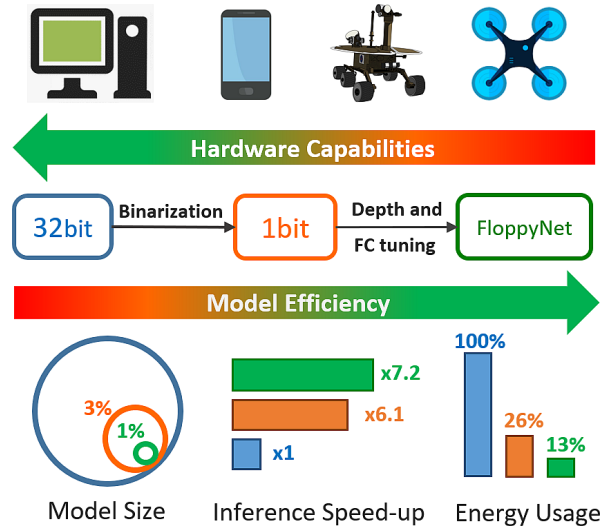}
\caption{FloppyNet is a compact and efficient binary network designed to enable VPR on edge devices and robots with severe hardware constraints.}
\label{fig:fig_1}
\end{figure}

Reducing resource demand while keeping VPR performance at a reasonable level is a difficult task. To tackle this challenge, in this paper we propose the multi-step approach summarized in Fig. \ref{fig:fig_1} that combines Binary Neural Networks (BNNs) \cite{courbariaux2016binarized, hubara2017quantized} and depth reduction to obtain very compact models that drastically decrease the memory requirements and improve computational efficiency. The subsequent VPR performance loss is mostly countered by training the model with a classifier stage including a reduced number of full-precision neurons. 

BNNs are a class of networks characterized by a single bit precision for both weights and activations instead of the 32 bits used by
conventional deep neural networks.
So far, BNNs have been employed and highly optimized  for classification tasks only, where they 
exhibit lower yet comparable accuracy to their full-precision counterparts
\cite{simons2019review,alizadeh2018a}. 
However, classification and VPR are different problems. The first aims to find the best fit among categories, while VPR consists of matching different images of the same scene. To the best of our knowledge, this paper is the first attempt to employ BNNs to solve the VPR problem effectively under environmental changes and with significantly reduced memory requirements and computational effort. Our best model\footnote{https://github.com/bferrarini/FloppyNet\_TRO}, dubbed FloppyNet, achieves comparable VPR performance to its full-precision and deeper counterparts while consuming $99\%$ less memory and running seven times faster. With a model size of 154 Kilobytes, FloppyNet can therefore be stored in an old \text{$5^{1/4}$-inch} floppy disk! 

The rest of this paper is organized as follows. Section \ref{sec:work} presents an overview of the related work. Section \ref{sec:bnnvpr} presents the proposed multi-step approach to obtain compact BNNs for VPR. The evaluation criteria and the experimental setup are described in Section \ref{sec:experiment}. A comprehensive binary layers analysis for VPR applications is proposed in Section \ref{ssec:layers}. VPR performance results and real-time benchmarks are presented and discussed in Sections \ref{ssec:comparison} and \ref{sec:deployment}, respectively. Section \ref{sec:handcrafted} compares the proposed BNN with several handcrafted image descriptors. Conclusions are drawn in Section \ref{sec:conclusions}.

\section{Related Work}
\label{sec:work}

\subsection{Visual Place Recognition}

Environmental changes, such as illumination and viewpoint variations, render VPR a very challenging task. As the core problem of VPR is image matching, computing a robust image representation is fundamental for developing reliable localization systems in dynamic environments.

Before CNNs became popular for computer vision applications, the techniques employed for image matching consisted mainly of handcrafted image descriptors. Histogram-of-Oriented Gradients (HOG) \cite{freeman1995orientation,dalal2005histograms} calculates the gradient of every image pixel to create a histogram, with each bar representing a gradient angle and magnitude. HOG can be either used for VPR as a global image descriptor \cite{mcmanus2014scene} or to describe regions of interest in an image \cite{zaffar2020cohog}. SIFT \cite{lowe2004distinctive} detects key-points in an image using Difference-of-Gaussian (DoG) and uses HOG to compute a descriptor of their neighborhood. 
SURF \cite{baya_speeded-up_2008} is partially inspired by SIFT. It is employed in a variety of VPR methods \cite{murillo2007surf,cummins2011appearance}.
Gist \cite{oliva2006building} is used for image matching in \cite{sunderhauf2011brief} and \cite{singh2010visual}. Gist employs a set of Gabor filters at different orientations and frequencies to extract global features from an image. Those features are then averaged and combined into a vector representing the image. CoHOG \cite{zaffar2020cohog} is a recent image descriptor proposed as a trainingless and computationally efficient alternative to CNN-based techniques. It uses image-entropy \cite{memorable_maps} to detect regions of interest that are subsequently assigned with a HOG descriptor. CoHOG is designed to achieve lateral viewpoint tolerance.

%
In recent years, machine learning techniques have become more and more popular in VPR applications. CNN-based methods achieve high performance in various environmental conditions \cite{zaffar2019levelling} and under viewpoint variations \cite{zaffar2019state}.
A pre-trained CNN for a different task can be used off-the-shelf for generating an image descriptor in place of handcrafted local and global image descriptors. 
For example, the features computed by the convolutional layers of AlexNet \cite{krizhevsky2012imagenet} can be used to match place images. How \etal{} \cite{7279659} showed that the features extracted from \textit{conv3} layer of AlexNet are robust to condition variations, while those from \textit{pool5} work well for viewpoint changes. 
Bai \etal{} \cite{bai2018sequence} used those layers' features to improve the matching performance of SeqSLAM \cite{milford2012seqslam} under viewpoint changes.  
AMOSNet and HybridNet \cite{chen2017deep} are variants of AlexNet trained on Specific PlacEs Dataset (SPED) \cite{chen2017deep} in order to compute more specific image representations for VPR. 
PlaceNet \cite{zhou2017places} is based on the same idea, but it uses VGG-16 \cite{Simonyan14c}, which is trained on a large dataset, dubbed Places365, organized in 365 place categories. CALC \cite{Merrill2018RSS} is a lightweight CNN proposed for addressing the loop closure detection problem efficiently. CALC is trained using an autoencoder to recreate a HOG descriptor from geometrically distorted place images. Cross-Region-Bow \cite{chen2017only}, the regional maximum activation of convolutions (R-MAC) \cite{tolias2016rmac}, CAMAL \cite{khaliq2019camal} and Region-VLAD \cite{khaliq2018holistic}, focus on features pooling from a pre-trained network as they consider feature extraction and aggregation as two separated stages. 
On the other hand, NetVLAD \cite{arandjelovic2016netvlad} consists of two stages trained end-to-end. The first is a VGG-16 network that extracts the features from an image followed by an aggregation layer to combine them in a VLAD-like descriptor \cite{jegou2010aggregating}. 

\subsection{Binary Neural Networks}

While CNNs are effective in addressing VPR, they include many parameters that result in a large model size and heavy computational effort. In the last decade or so, several techniques have been proposed to decrease models’ runtime requirements. 
Early approaches targeted redundant and non-informative weights: Optimal Brain Damage \cite{lecun1990optimal} and Optimal Brain Surgeon \cite{hassibi1992second} decrease the number of connections using the Hessian of the loss function. Han \etal{},  \cite{han2015learning} showed how to reduce the number of parameters by one order of magnitude in several \art{} networks by weight pruning. 
Also, a network's size can be shrunk by lowering the precision of the weights. However, post-quantization yields performance loss, which is more prominent as the precision lowers. In particular, post-training binarization (1-bit precision) enables the highest model compression and computational speed-up but impacts heavily on a classifier's accuracy \cite{courbariaux2014training}.

Binary-aware training enables low precision models with acceptable classification accuracy \cite{simons2019review}.
Although training binary models from scratch was attempted decades ago \cite{saad1990training}, only recently, gradient-based techniques have become applicable to BNNs. 
Courbariaux \etal{} \cite{courbariaux2016binarized} trained a full binary network for the first time using Straight-Through-Estimator (STE) \cite{bengio2013estimating}. 
The key idea of STE is to keep in memory real-valued weights, which are binarized only in the forward pass to compute neurons' activation and updated during back-propagation as in a standard neural network. 
Afterwards, several additions to the field were proposed to improve BNNs. In XNOR-Net \cite{rastegari2016xnor},
the convolutional blocks are rearranged to increase classification accuracy. Batch-Normalization (BatchNorm) is usually placed after the convolution and before the activation function.
In XNOR-Net, BatchNorm and binary activation precede convolution so that pooling occurs before binarization.
DoReFa-Net \cite{zhou2016dorefa} exploits bitwise operations to compute the dot product between a layer's weights and the inputs in an efficient way to speed-up training. 
In \cite{esser2019learned}, binarization threshold is learned along with the weights to shorten the accuracy gap with full-precision classifiers. 
Networks using a less extreme quantization have been proposed as a more accurate alternative to binary networks. Ternary networks \cite{li2016ternary,zhu2016trained} use three values to encode weights.
Although they exhibit a significant memory reduction and simple arithmetics, ternary networks require 2-bits to store weights and do not outperform BNNs by a wide margin \cite{simons2019review}.

To the best of our knowledge, BNNs have been used only for classification so far. Unlike regular CNNs, BNNs have not been considered for VPR yet. This paper aims to contribute to the field by proposing a highly compact class of binary networks to solve the VPR problem effectively in changing environments.


\subsection{BNNs Inference Frameworks}
\label{ssec:compute_engine}

Binarization reduces a model's size dramatically and enables efficient convolution computation. However, BNNs cannot express their full potential without an inference engine that can efficiently compute bitwise operations.
Different hardware platforms implement binary primitives, such as XNOR and pop-count, differently. Therefore, inference libraries typically target one or a few hardware platforms to guarantee that the deployed models can run efficiently. In this section, we present a selection of the most relevant libraries and tools for deploying BNNs.

Courbariaux \etal{} \cite{courbariaux2016binarized} released a GPU kernel-based that speeds up convolutions by seven times using SIMD (single instruction, multiple data) within a register (SWAR) technique. 
DaBNN \cite{cai2019once} is a stand-alone library to deploy BNNs on ARM platforms. Binary convolutions are computed by combining im2col transformation and GEMM (GEneral Matrix Multiplication). DaBNN uses ad hoc implementation written in ARM assembly that enables $8-10\times$ speed-up compared to a full-precision implementation.
BMXNet \cite{zhuang2019structured} extends MXNet \cite{chen2015mxnet} providing a complete ecosystem to train and deploy BNNs. Like DaBNN, it computes binary convolutions by combining im2col with GEMM. BMXNet utilizes standard C++ to implement binary operations reaching a $13\times$ speed-up compared to floating-point convolutions written with the CBLAS library \cite{whaley2005minimizing}.
Riptide \cite{MLSYS2020_2a79ea27} is another end-to-end framework for training and deploying BNNs, focusing on integrating binary convolutions with those layers that cannot be binarized, such as BatchNorm and activation functions. Models are trained with Tensorflow \cite{tensorflow2015-whitepaper} and compiled for deployment with TVM \cite{chen2018tvm}. A binary model compiled with Riptide is $4\times$ to $12\times$ faster than its full-precision counterpart.
Larq Compute Engine (LCE) \cite{bannink2020larq} is part of the Larq project \cite{larq} to train and deploy BNNs.
As Riptide, LCE aims to integrate efficiently binary convolutions with a model's full-precision components. Also, LCE proposes a binary kernel highly optimized for ARM platforms. The convolution speed-up is $8.5-18.5\times$.
Finally, FINN \cite{umuroglu2017finn} is an experimental tool targeting FPGAs supporting PYNQ \cite{pynq.io}. FINN does not include a training framework but can compile PyTorch \cite{NEURIPS2019_9015} models optimized with Bravitas \cite{brevitas}.

\subsection{Deep Neural Networks Benchmark Analysis}
\label{ssec:bench}

Benchmark analysis is an important step to take for understating a model's usability with the target hardware.
The dominant evaluation metrics are inference time (or inference latency), power consumption and memory usage. Those metrics are particularly relevant for robotic applications as they often rely on constrained hardware and battery supply. TANGO \cite{8695662} employs those metrics to assess CNN models deployed on several hardware platforms. The importance of energy usage is emphasized by Palit \etal{} \cite{8942095}, who present an energy estimation model along with empirical data from well-established CNNs. DNNTune \cite{xia2019dnntune} uses inference time and energy consumption to tune both CNNs and quantized networks for several application scenarios.

Finding the best possible trade-off between memory usage and performance is an essential task when resource-constrained hardware is employed with deep neural networks.
Howard \etal{} \cite{HowardZCKWWAA17} proposed a class of efficient networks for mobile applications using the classification accuracy to parameters number as a tuning criterion.
Bianco \etal{} \cite{8506339} employs the \textit{accuracy density} metric to represent the efficiency of deployed models at using their parameters. 
Unlike CNN, BNN's memory allocation efficiency received little or no attention so far. BNNs' memory footprint is usually evaluated only relatively to full-precision counterparts disjointly from their performance \cite{rastegari2016xnor,bethge2019simplicity}. To fill this gap, this paper extends the accuracy-parameter trade-off analysis to quantized networks and proposes a metric to assess the memory allocation efficiency of BNNs for VPR applications.

\begin{figure}[pt]
\centering
\includegraphics[width=8.2cm]{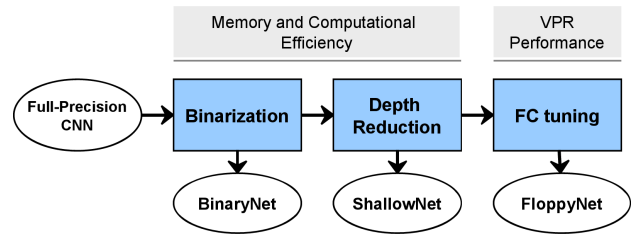}
\caption{The proposed approach consists of three steps: Binarization reduces the model size by about 97\%. Depth reduction decreases the number of layers for further model size and MAC number reduction. The subsequent performance loss due to binarization is mainly countered by training the network with an appropriately sized fully-connected stage consisting of full-precision neurons.}
\label{fig:steps}
\end{figure}

\section{Binary Neural Networks for VPR}
\label{sec:bnnvpr}

This section presents a new class of binary neural networks for visual place recognition and provides implementation details. To achieve memory and computational efficiency while maintaining reasonable VPR performance, we propose a multi-step approach to turn a standard CNN into a compact yet effective feature extractor.
Fig. \ref{fig:steps} summarizes the process. Binarization reduces the model size and speeds up convolutions by enabling bitwise operations. Depth reduction decreases the number of layers for further model size reduction and faster computation. The subsequent performance loss due to binarization and layer removal is countered chiefly by training the network with an appropriately sized fully-connected stage consisting of full-precision neurons.

\subsection{Binarization}
\label{ssec:binarization_step}

binarization improves both memory usage and computational speed-up. 
Storing a 32-bit weight requires four bytes, while a single bit is needed for a binary one. Hence, by concatenating $32$ binary weights into a floating-point variable, the resulting model size is about $97\%$ smaller than its full-precision counterpart.
Similarly, bitwise operations between weights and activations are computed in parallel with $32$ binary operands resulting in a significant convolutions speed-up.

Training a binary model with a reasonable performance gap from its full-precision counterpart requires applying specific techniques and some network structure adjustments. 
This section has the two-fold purpose of describing the implementation criteria we have taken and giving a gentle introduction to BNNs.

\begin{figure}[t]
\centering
\includegraphics[width=\halfwidth{}]{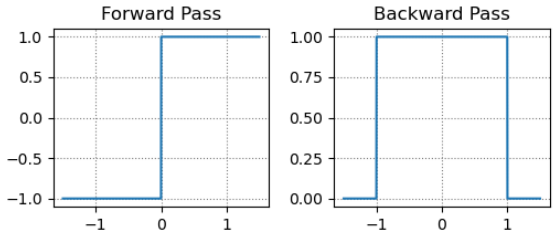}
\caption{Sign quantizer in forward and backward passes.}
\label{fig:quantizers}
\end{figure}

\subsubsection{Training and Binary Function}

training BNNs with backpropagation is not applicable as it requires a sufficient precision to allow gradient accumulation to work \cite{hubara2017quantized}.
Courbariaux \etal{} solved this problem \cite{courbariaux2016binarized} with Straight Through Estimator (STE) \cite{bengio2013estimating}. 
The fundamental idea of STE is that the quantization function is applied in the forward pass but skipped during backpropagation.
STE keeps a set of full-precision weights denoted as proxies ($W_F$) which are binarized ($W_{B}$) on forward pass to make a prediction and compute a loss. Any function can be used as binarization function. 
Courbariaux \etal{} use $sign$ function:
\begin{equation}
    W_{B} = sign(W_F)
\end{equation}
In the backpropagation phase, $W_F$ are updated accordingly to the loss gradient as in a regular network:
\begin{equation}
    \frac{\partial L_{oss}}{\partial W_{F}} = \frac{\partial L_{oss}}{\partial W_{B}}
\end{equation}
Activations are binarized in the forward pass similarly to the weights. Fig. \ref{fig:quantizers} shows the plots for the binarization function. In the forward pass, it behaves as the $sign$ function performing binarization. In the backward pass the function returns a clipped identity of the gradient. Courbariaux \etal{} \cite{courbariaux2016binarized} observed that canceling the gradient when activation exceeds $1.0$ improves a model's accuracy.

\begin{equation}
  \frac{\partial L_{oss}}{\partial a_{F}} =\left\{
  \begin{array}{@{}ll@{}}
    \frac{\partial L_{oss}}{\partial W_{B}}, &  \text{if $|a_{F}| \le 1$} \\
    0, & \text{otherwise}
  \end{array}\right.
\end{equation}
The binary models presented in this work use $sign$ as a quantizer and are trained with Larq \cite{larq}. Larq is a framework built on top of Keras \cite{chollet2015keras} which offers full support to train BNNs with STE.

%
\subsubsection{Encoding Values}
\label{ssec:encoding}

Binary encoding of weights and activations reduces dot products to a series of bitwise operations. In particular, representing logical `0` and `1` with $-1$ and $1$ renders convolutions and matrix multiplications a series of XNOR and pop-count operations \cite{courbariaux2016binarized}.
However, to exploit the efficiency of binary operations, a dedicated compute engine or specific hardware is required \cite{courbariaux2016binarized,hubara2017quantized}.
A conventional compute engine
stores binary weights into 32-bit variables. As a result, multiply-accumulate operations (MAC) in BNNs require the same time and resources as in a full-precision network.  
A proper compute engine concatenates 32 binary variables into a 32-bit register and evaluate them altogether using bitwise operations. 
Typically, a binary MAC is implemented as follows: 
\begin{equation}
    a_1 \mathrel{+}= \text{popcount}(\text{xnor}(a_o^{32},w_1^{32}))
\end{equation}
where $a_o^{32}$ and $w_1^{32}$ are sets of 32 inputs and weights. Although weights concatenation enables the computation of multiple binary MACs in parallel, $32\times$ speed-up is unrealistic. This limitation depends on several factors, including instruction scheduling, CPU pipeline stalls, and the hardware instruction set above anything else. General purpose CPUs and GPUs have specialized instructions for fusing a floating-point MAC in a single clock cycle. Conversely, in the binary case, no such instructions exist. Hence, a binary MAC results from multiple instructions on many hardware platforms such as Nvidia GPUs \cite{cuda_thoughput} and ARM processors \cite{arm_thoughput}. Therefore, if we let $c_b$ represents the number of clock cycles to compute a binary MAC for a given hardware platform, then the obtainable speed-up is capped at $32/c_b$.
\begin{figure}[t]
\centering
\includegraphics[width=\halfwidth]{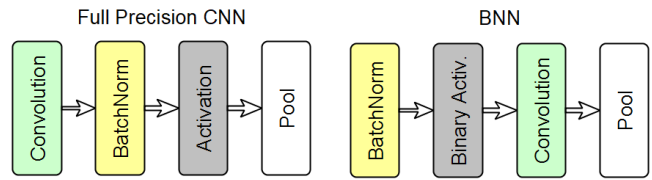}
\caption{Convolutional blocks in a CNN (left) and in a BNN (right).}
\label{fig:batchnorm_order}
\end{figure}
%

\begin{figure*}[!ht]
    \centering
    \includegraphics[width=14.5cm]{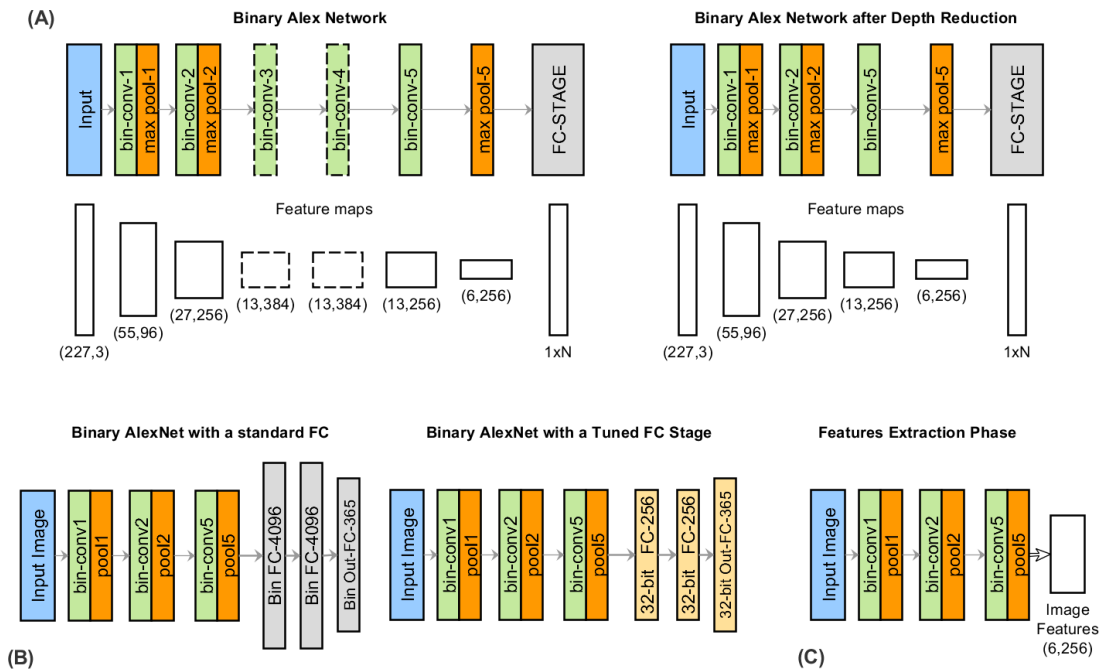}
    \caption{Depth Reduction (A) and FC tuning (B) applied to AlexNet. Depth reduction consists of removing \textit{conv3} and \textit{conv4} layers. The three pooling layers are kept to maintain the exact shape of the \textit{pool5} output feature map (c).}
    \label{fig:depth_reduction}
\end{figure*} 

\subsubsection{Batch Normalization}
\label{ssec:batch-norm}

Batch Normalization (BatchNorm) \cite{batchnorm} uses mini-batch statistics during training to adjust and scale activations. The central role of BatchNorm in full-precision networks is to speed up the training. In BNN, BatchNorm is essential as it improves performance and helps training convergence \cite{alizadeh2018a,Sari2019HowDB,simons2019review}.
It is worth mentioning that the parameters of BatchNorm layers cannot be binarized; however, they are few compared with the number of weights and do not contribute significantly to a model's size (Table \ref{tab:memory}).

\subsubsection{Layers Order}
\label{ssec:pooling}

a convolutional block in a CNN consists of convolution, BatchNorm, activation and pool. BNNs achieve better performance if the order of the layers is as follows: BatchNorm, binary activation, convolution and pool \cite{rastegari2016xnor}.
This layer arrangement has a two-fold purpose. 
First, it allows for pooling from real values before binarization. 
Otherwise, the result would be a tensor dense in 'ones' which is proven to negatively affect the accuracy of a BNN \cite{alizadeh2018a}. 
Second, BatchNorm can replace bias as it works as a threshold for the subsequent layer \cite{Sari2019HowDB,simons2019review}. 
As bias parameters cannot be binarized, not using them reduces the memory and the number of full-precision MACs in binary networks. The BNNs implemented for this work do not use bias but use BatchNorm modules instead.

In the rest of this paper, the term convolutional block implies the presence of BatchNorm in the proper position as shown in Fig. \ref{fig:batchnorm_order}.

\subsubsection{First Layer Input}
\label{ssec:first_layer}

full-precision inputs are recommended to improve a binary model's accuracy \cite{hubara2017quantized}. The model size is unaffected since the weights are binary and the impact of computational speed is acceptable when the convolution filters is few compared to the deeper layers.
Accordingly with this consideration, the binary networks presented in this work have the first convolutional layer directly
connected to the input image with no binary activation and
BatchNorm placed in the middle.

\subsubsection{Padding}
\label{ssec:padding}

in full-precision networks, convolutions are often padded with zeros. This standard practice cannot be applied to BNNs that require padded values within the encoding set to enable bitwise operations. Zero-padding, in a BNN using $\{-1,1\}$, adds a third value along with $-1$ and $1$ that renders convolutions incompatible with bitwise operations.
Our models, therefore, use one-padding accordingly with the weights encoding $\{-1,1\}$.

\begin{figure}[t!]
    \centering
    \includegraphics[width=6.2cm]{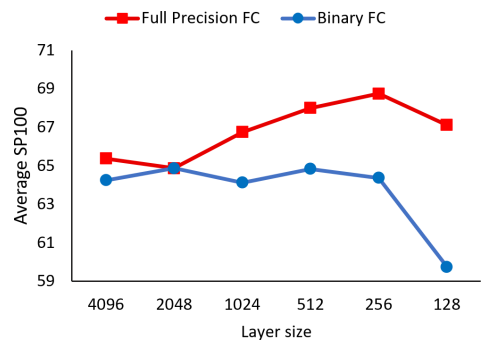}
    \caption{Average $\ratio{}$ across all the test datasets for full-precision and binary fully-connected stages at different layer sizes.}
    \label{fig:fc_study}
\end{figure}

\subsection{Depth Reduction}
\label{ssec:depth_reduction}

The primary motivation for depth reduction is to decrease the number of a model's parameters. Networks for classification are deep and can have dozens of convolutional levels \cite{khan2020survey}. 
However, VPR is a different task and we empirically found that it is possible to achieve good performance with fewer layers and weights. 
Not only the model size but also the computational efficiency of the network benefits from depth reduction. For example, our best model is obtained by removing the two intermediate convolutional layers from an AlexNet-like CNN 
as shown in Fig. \ref{fig:depth_reduction}.A. 
This operation decreases by $66\%$ the weights amount yielding significant model size and MACs reduction (Table \ref{tab:memory_floppy}). In our experiments, the network resulting from depth reduction is denoted by ShallowNet (Fig. \ref{fig:steps}).

\subsection{Fully-Connected Stage Tuning}
\label{ssec:fc_tuning}

BNNs are highly optimized for classification. The fully-connected (FC) stage of classifiers is often populated with a large number of neurons. AlexNet and VGG16, for example, include 4096 units in each layer. 
When it comes to training a model for VPR, the hyper-parameters of the FC layers should be revised. Our best binary model is trained using 256 32-bit neurons per FC layer (Fig. \ref{fig:depth_reduction}.B). The optimal FC size of 256 neurons has been determined empirically by training and testing several models. The use of a full-precision instead of binary FC stage is based on the following considerations. 
Binary weights are a source of gradient noise \cite{hubara2017quantized} that renders the training more complex and longer to complete \cite{toneva2018an}. 
Using 32-bit FC reduces the number of binary weights that need to be learnt, making training more stable. Smoother training has a lower chance to overshoot a loss function’s minimum resulting in a better optimized model. 
Fig. \ref{fig:fc_study} shows the VPR performance of the proposed model for various FC sizes. The performance peak corresponds to 256 full-precision neurons. It is relevant mentioning that FC stage tuning is applicable only when VPR is carried out with convolutional features (Fig. \ref{fig:depth_reduction}.C).This is the case of the proposed FloppyNet, which uses \textit{pool5} features for VPR, as detailed in Section \ref{ssec:floppynet}.

\section{Experimental Setup}
\label{sec:experiment}

This section provides details about the experimental setup (including evaluation criteria, training and test datasets) used for assessing the VPR performance of the binary neural networks presented in Section \ref{ssec:networks}.

\subsubsection{VPR Performance}

visual palce recognition is cast as a loop closure detection task \cite{bai2018sequence}. Reference images showing already visited locations are searched to find the best match with the robot camera's current view, namely the query image. 
VPR is considered successful when a query image is paired with one of the correct reference images.
The image descriptors used to match images are obtained by L2-normalization of a network's layer output:
\begin{equation}
    D = \frac{\hat{X_l}}{||\hat{X_l}||_2}\text{,}
\end{equation}
where $\hat{X_l}$ is the output of the $l^{th}$ layer.

Descriptors are compared using Euclidean distance; the shorter the distance, the higher the similarity between two images.
\begin{equation}
    d = ||D_1 - D_2||_2{,}
    \label{eq:dist}
\end{equation}
where $D_1$ and $D_2$ are the image descriptors to be compared.
The reference image with the shortest distance from the query is regarded as the current location.

Following the approach proposed in \cite{EP}, VPR is evaluated on a whole dataset
with $\ratio{}$ index. It represents the 
ratio 
of places that are correctly recognized against the ground truth. 

\subsubsection{Memory Allocation Efficiency}
\label{ssec:mem_eff}

VPR performance is also evaluated in relation to memory requirements. We define memory efficiency as the ratio of the model size to $\ratio{}$:
\begin{equation}
    \meff{} = \frac{M_{size}}{\ratio{}}
    \label{eq:efficiency}
\end{equation}

$\meff{}$ measures the memory cost per $\ratio{}$ point, expressing the trade-off between memory usage and VPR performance. 
The lower is $\meff{}$, and the more efficient is the model at using memory. 
Memory efficiency is a generalization of the trade-off analysis between accuracy and parameters density \cite{8506339} to low-precision networks whose memory footprint also depends on weight quantization. Hence, the use of model size instead of the number of weights in Eq. \ref{eq:efficiency}. 
Moreover, $\meff{}$ can be applied to networks having the same structure but different weight precision to determine the relationship between VPR performance and quantization, providing additional information to characterize a low precision network or choose the optimal quantization for an application.

\begin{figure}[!t]
    \centering
    \includegraphics[width=\halfwidth{}]{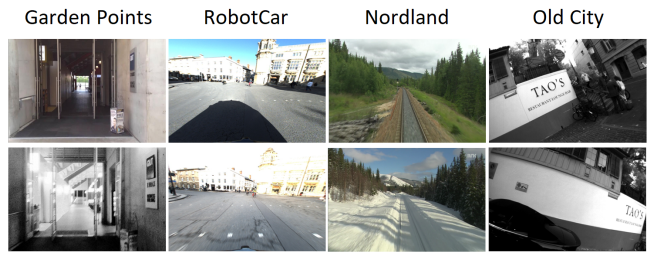}
    \caption{A corresponding image pair from each test dataset.}
    \label{fig:dataset_Example}
\end{figure} 

\begin{table}[b]
\caption{Test datasets and ground truth tolerance.}
  \begin{adjustbox}{width=\columnwidth,center}
    \begin{tabular}{lcccc}
    \multicolumn{1}{c}{\multirow{2}[1]{*}{\textbf{Dataset}}} & \multicolumn{1}{c}{\multirow{2}[1]{*}{\textbf{Viewpoint}}} & \multicolumn{1}{c}{\multirow{2}[1]{*}{\textbf{Conditional}}} & \textbf{Reference} & \textbf{Ground} \\
          &       &       & \textbf{Images} & \textbf{Truth} \bigstrut[b]\\
    \hline
    \hline
    GardenPoints & Lateral & Day-Night & 200   & 2 frames \bigstrut\\
    \hline
    \multirow{2}[2]{*}{Nordland} & \multirow{2}[2]{*}{None} & Summer & \multirow{2}[2]{*}{309} & \multirow{2}[2]{*}{5 frames} \bigstrut[t]\\
          &       & Winter &       &  \bigstrut[b]\\
    \hline
    Old City & 6-DOF & None  & 6708  & by authors \bigstrut\\
    \hline
    RobotCar & \multirow{2}[2]{*}{Lateral} & Dinamic Obj & \multirow{2}[2]{*}{206} & \multirow{2}[2]{*}{5 frames} \bigstrut[t]\\
         Cross-Seasons &       & Illumination &       &  \bigstrut[b]\\
    \hline
    \end{tabular}%
    \label{tab:dataset}%
  \end{adjustbox}
\end{table}%

\begin{table*}[tp]
  \caption{Baseline structure and model size before and after binarization. Size Ratio is between binarized and full-precision models.}
  \begin{adjustbox}{width=\textwidth,center}
    \begin{tabular}{lcccccccccc}
    \multicolumn{11}{c}{\textbf{Baseline Architecture}} \bigstrut\\
    \hline
    \hline
          & \textbf{conv1} & \textbf{pool1} & \textbf{conv2} & \textbf{pool2} & \textbf{conv3} & \textbf{conv4} & \textbf{conv5} & \textbf{pool5} & \textbf{fc6} & \textbf{fc7} \bigstrut\\
    \hline
    Layer Setup & C(11,4,96) & P(2,2) & C(5,1,256) & P(2,2) & C(3,1,384) & C(3,1,384) & C(3,1,256) & P(2,2) & FC(4096) & FC(4096) \bigstrut[t]\\
    Features Size & 290400 & 290400 & 186624 & 43264 & 64896 & 64896 & 43264 & 9216  & 4096  & 4096 \\
    Parameters (M) & 0.03  & 0.04  & 0.65  & 0.65  & 1.54  & 2.86  & 3.75  & 3.75  & 41.5  & 58.29 \\
    Model Size (KiB) & 136.5  & 137.3  & 2538  & 2540  & 5998  & 11186 & 14646  & 14648  & 162120 & 227704 \\
    Total MACs (M) & 105.42 & 105.42 & 553.31 & 553.31 & 702.83 & 927.11 & 1076.63 & 1076.63 & 1114.38 & 1131.16 \bigstrut[b]\\
    \hline
    Binarizable Par. (M) & 0.03  & 0.03  & 0.65  & 0.65  & 1.53  & 2.86  & 3.75  & 3.75  & 41.49 & 58.27 \bigstrut[t]\\
    Non-Binarizable Par. & 0     & 0     & 192   & 192   & 704   & 1472  & 2240  & 2240  & 2752  & 10944 \\
    Binary Model Size (KiB) & 4.25  & 4.25  & 80    & 80    & 190   & 355   & 466   & 466   & 5076  & 7156 \\
    Model Size Ratio (\%) & 3.12  & 3.1   & 3.15  & 3.15  & 3.17  & 3.17  & 3.18  & 3.18  & 3.13  & 3.14 \bigstrut[b]\\
    \hline
    \end{tabular}%
    \end{adjustbox}
  \label{tab:memory}%
\end{table*}%

\begin{table*}[ht]
  \centering
  \caption{Proposed BNNs structure and model size.}
 
    \begin{tabular}{lcccccc}
    \multicolumn{7}{c}{\textbf{FloppyNet and ShallowNet Fetures Extractors}} \bigstrut[b]\\
    \hline
    \hline
          & \textbf{conv1} & \textbf{pool1} & \textbf{conv2} & \textbf{pool2} & \textbf{conv5} & \textbf{pool5} \bigstrut\\
    \hline
    Layer Setup & C(11,4,96) & P(2,2) & C(5,1,256) & P(2,2) & C(3,1,256) & P(2,2) \bigstrut[t]\\
    1-bit parameters (M) & 0.03  & 0.03  & 0.65  & 0.65  & 1.24  & 1.24 \\
    32-bits parameters (M) & 0     & 0     & 192   & 192   & 704   & 704 \\
    Model Size (KiB) & 4.25  & 4.25  & 80    & 80    & 154   & 154 \\
    Total MACs (M) & 105.42 & 105.42 & 553.31 & 553.31 & 652.99 & 652.99 \bigstrut[b]\\
    \hline
    Parameter Rate \% & 100   & 100   & 100   & 100   & 33.1  & 33.1 \bigstrut[t]\\
    Size Rate \% (BinaryNet) & 100   & 100   & 100   & 100   & 33.05 & 33.05 \\
    Size Rate \% (Baseline) & 3.12  & 3.1   & 3.15  & 3.15  & 1.05  & 1.05 \\
    MAC Rate \% & 100   & 100   & 100   & 100   & 60.7  & 60.7 \bigstrut[b]\\
    \hline
    \end{tabular}%
  \label{tab:memory_floppy}%
\end{table*}%

\subsubsection{Inference Latency and Power Usage}
\label{ssec:deply_metrics}
\noindent inference latency,  $\tf{}$, and power usage ($\power{}$) measurments are taken from deployed models.
$\tf{}$ is the time required by a model to complete a forward pass. The time intervals to load an image and consume the output are excluded from the measurement so that $\tf{}$ reflects the actual
computational 
effort for an image representation. 
$\power{}$ is measured directly on the hardware platform and used to determine the inference energy cost: 

\begin{equation}
    \energy{} = \power{}\tf{}\text{.}
    \label{eq:energy}
\end{equation}
%
%

\subsection{Training Data}
\label{ssec:TR_data}

The dataset used to train all the models is Places365 \cite{zhou2017places,places365-gith}. It is a place-themed dataset consisting of 1,803,460 images divided into 365 categories with between 3068 and 5000 images in each category. The validation set includes 100 images per location class.

\subsection{Test Data}
\label{ssec:TE_data}

On long term runs, a robot visits a place at different times or from different directions. These factors yield changes in the appearance of places captured by the robot's camera. In order to provide comprehensive results, test data includes four datasets, each containing environmental and/or viewpoint changes. 
All datasets have two subsets that correspond to different traverses of the environment (Fig. \ref{fig:dataset_Example}). One is the reference dataset representing the previous knowledge of the environment while the other is the query dataset that represents the current traverse.
The datasets include a different number of query images. To compute fair average performance indicators (e.g. Table \ref{tab:baseline_features}), we randomly sampled 200 query images from each of them for a total of 800 images. 

The datasets are detailed below and summarized in Table \ref{tab:dataset}.


\subsubsection{GardenPoints Walking \cite{7353986}}
\label{sss:garden}

this dataset includes three traverses of the Queensland University of Technology (QUT). The experiments employed Right-Day and Right-Night to test VPR under illumination changes and mild lateral shifts. Ground truth is built by frame correspondences with a tolerance of $\pm{}2$ frames \cite{lowry2016supervised}.

\subsubsection{Nordland \cite{nordlands}}
\label{sss:nordland}

a set of four traverses captured along a rail track in Norway in every season. The experiment employed Summer and Winter journeys as reference and query datasets, respectively. The ground truth is built with a tolerance of $\pm{}5$ frames \cite{lowry2016supervised}.

%

\subsubsection{Old City \cite{maffra2019real} }
\label{sss:old_city}

a urban dataset with two traverses showing the same location from different perspectives to generate the viewpoint variations experienced by a 6-DOF (Degrees-Of-Freedom) aerial robot. The ground truth data is available from the authors \cite{V4RLdata}.

\subsubsection{RobotCar Cross-Seasons \cite{larsson2019cross}}
\label{sss:sped200}
a subset of the Oxford RobotCar dataset \cite{RobotCarDatasetIJRR} consisting of two sequences of 206 sunny query images and 202 dusk reference images recorded on board of a car driving in an urban environment. 
RobotCar dataset includes illumination changes, mild lateral viewpoint shifts and dynamic elements such as pedestrians, cars and shadows. 
Ground truth is built by frame correspondences with a tolerance of $\pm{}5$ frames.

\begin{figure}[!t]
    \centering
    \includegraphics[width=\halfwidth{}]{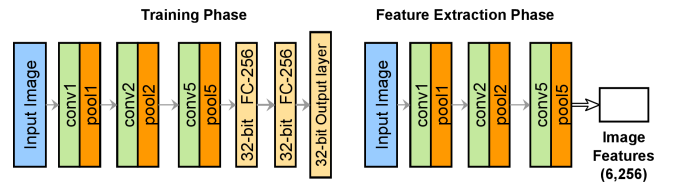}
    \caption{FloppyNet layer structure. 
    The output features used for VPR are from the \textit{pool5} layer. 
    }
    \label{fig:floppynet}
\end{figure}

\begin{table*}[!t]
  \centering
  \caption{$S_{p100}$ for every layer in Baseline.}
    \begin{tabular}{lcccccccccc}
    \multicolumn{11}{c}{\textbf{Baseline  Image  Features}} \bigstrut[b]\\
    \hline
    \hline
          & \textbf{conv1} & \textbf{pool1} & \textbf{conv2} & \textbf{pool2} & \textbf{conv3} & \textbf{conv4} & \textbf{conv5} & \textbf{pool5} & \textbf{fc6} & \textbf{fc7} \bigstrut\\
    \hline
    Garden Point & 14.5  & 29    & 56.5  & 62.5  & 67    & 79.5  & 66    & 84    & 73    & 56.5 \bigstrut[t]\\
    RobotCar & 80.1  & 89.3  & 87.4  & 90.3  & 91.3  & 92.2  & 82    & 93.2  & 90.8  & 74.3 \\
    Nordland & 69    & 75.5  & 86.5  & 91    & 91    & 95    & 54    & 85    & 40    & 29 \\
    Old City & 7     & 11    & 7     & 9     & 11    & 18    & 13    & 33    & 39    & 37 \\
    \hline\rule{0pt}{2.6ex}
    Average & 42.7  & 51.2  & 59.4  & 63.2  & 65.1  & 71.2  & 53.8  & \textbf{73.8} & 60.7  & 49.2 \\
    \hline
    \end{tabular}%
  \label{tab:baseline_features}%
\end{table*}%

\begin{table*}[!t]
  \centering
  \caption{$S_{p100}$ for every layer in BinaryNet.}
    \begin{tabular}{lcccccccccc}
    \multicolumn{11}{c}{\textbf{BinaryNet  Image  Features}} \bigstrut[b]\\
    \hline
    \hline
          & \textbf{conv1} & \textbf{pool1} & \textbf{conv2} & \textbf{pool2} & \textbf{conv3} & \textbf{conv4} & \textbf{conv5} & \textbf{pool5} & \textbf{fc6} & \textbf{fc7} \bigstrut\\
    \hline
    Garden Point & 3     & 11.5  & 43    & 51.5  & 75.5  & 74.5  & 76    & 79.5  & 66.5  & 39.5 \bigstrut[t]\\
    RobotCar & 61.2  & 73.3  & 81.4  & 81.9  & 82.4  & 82.4  & 83    & 83    & 84    & 63.6 \\
    Nordland & 53    & 60    & 71.5  & 71    & 77.5  & 77    & 72    & 71.5  & 38.5  & 14 \\
    Old City & 9     & 10.5  & 11.5  & 15    & 16    & 15    & 20.5  & 23    & 49    & 44.5 \\
    \hline\rule{0pt}{2.6ex}
    Average & 31.6  & 38.8  & 51.9  & 54.9  & 62.9  & 62.2  & 62.9  & \textbf{64.3} & 59.5  & 40.4\\
    \hline
    \end{tabular}%
  \label{tab:binarynet_features}%
\end{table*}%

\subsection{Binary Network and Comparison Baseline}
\label{ssec:networks}

The networks used for the experiments are based on the AlexNet archetype \cite{krizhevsky2012imagenet}, which is one of the most used network types for VPR \cite{chen2017only,tolias2016rmac,arandjelovic2016netvlad,khaliq2018holistic}.
AlexNet-type networks' structure consists of several convolutional blocks (CB) alternated with pool layers followed by a fully-connected (FC) stage with one or more hidden layers. 
 
The baseline CNN, denoted by Baseline in this paper, is very similar to a standard AlexNet \cite{krizhevsky2012imagenet} except for the use of BatchNorm and pool layers with a $2 \times 2$ non-overlapping kernel for higher accuracy \cite{krizhevsky2012imagenet}. 
Baseline network has five convolutional blocks followed by a fully-connected stage with two hidden layers, including 4096 neurons each. The detailed structure is shown in Table \ref{tab:memory} using the following notation. $C(k,s,h)$ indicates a convolutional block with kernel size $k$, stride $s$ and $h$ channels (filters). A similar notation is used for max pooling: $P(k,s)$. Fully-connected layers are indicated with $FC(n)$, where $n$ is number of neurons. 
The model sizes and MACs reported in Table \ref{tab:memory} are cumulative per network layer. For example, if the baseline is cut to use \textit{fc6} features, the corresponding size of the model is $158.32$ MiB and the MACs are $1.1$ billion. 
BinaryNet is the binary version of Baseline.
The bottom part of Table \ref{tab:memory} shows the number of binarizable parameters. The remaining 32-bit parameters are due to BatchNorm as described in Section \ref{ssec:batch-norm}. However, their contribution to the binary model size is negligible. BinaryNet sizes vary from the $3.12\%$ of Baseline at $conv1$, which is not preceded by a BatchNorm layer, to $3.18\%$ at $pool5$.

\begin{figure*}[!ht]
	\vspace*{1ex}
	\centering
	\textsc{\small\figtiTFlitefont{VPR matching performance comparison}}\par\vspace*{3ex}
		\begin{annotatedFigure}{\includegraphics[width=0.32\linewidth]{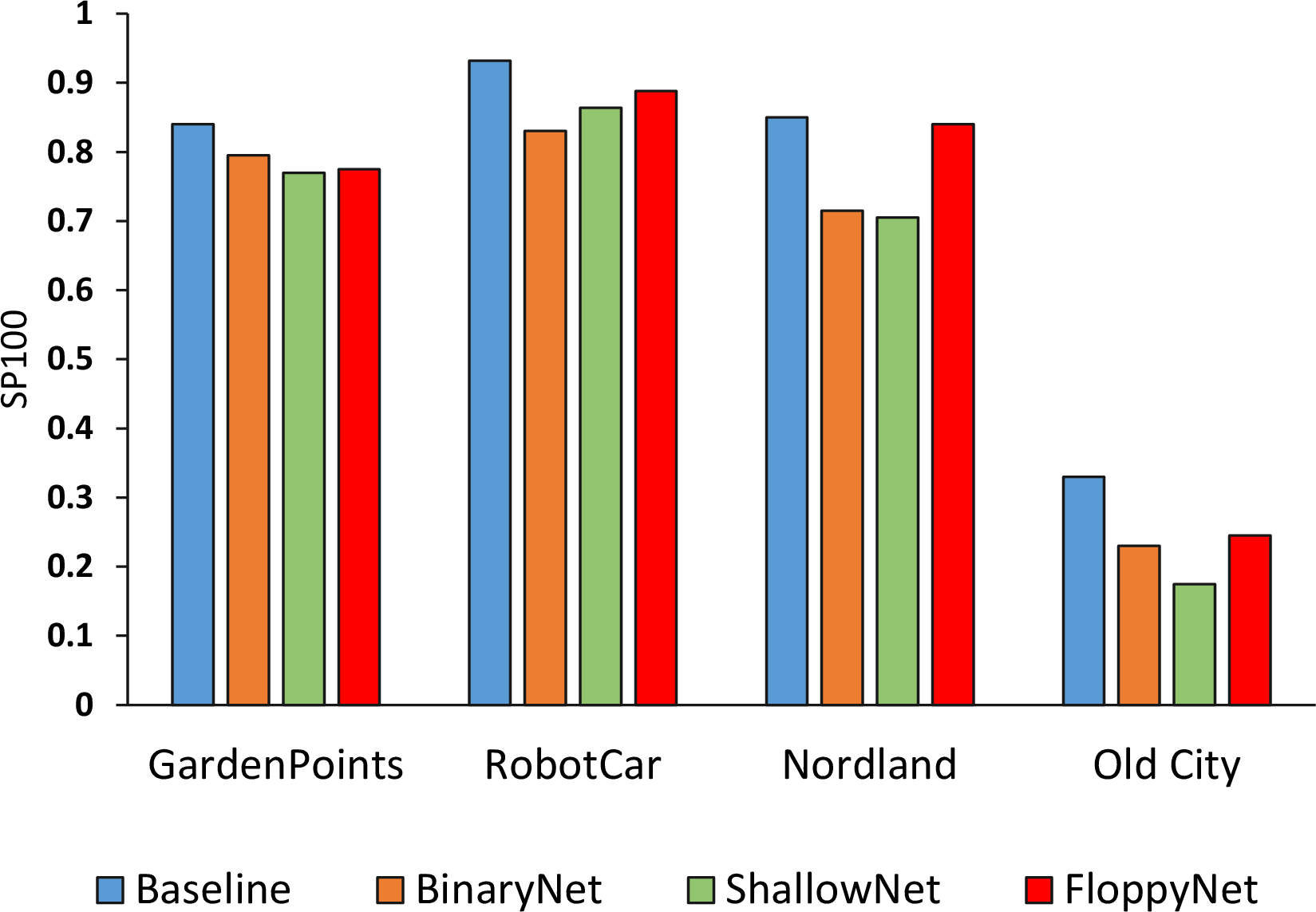}}
			\annotatedFigureText{0.90,0.95}{black}{0.00}{\small{(A)}}
		\end{annotatedFigure}
		\begin{annotatedFigure}{\includegraphics[width=0.32\linewidth]{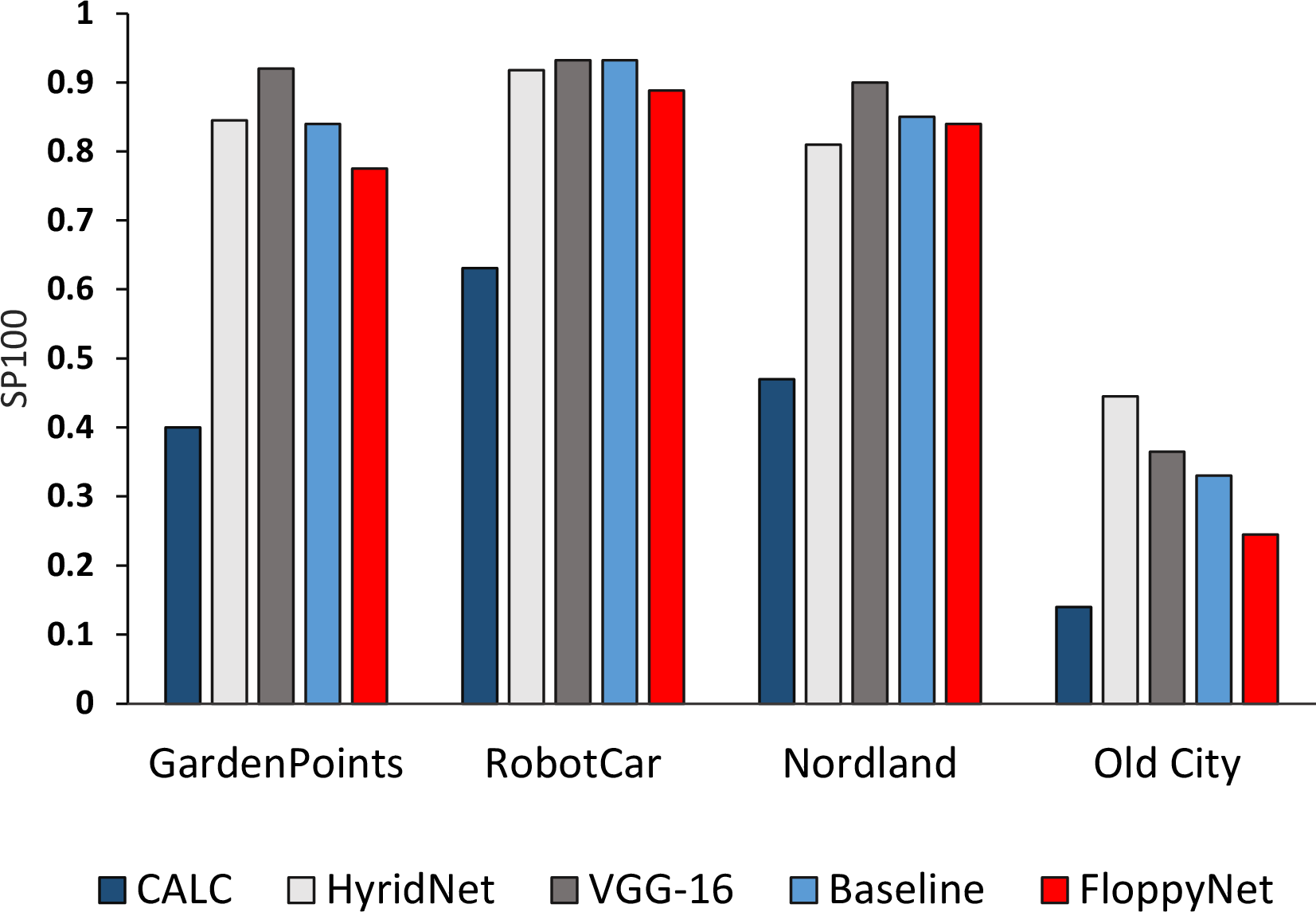}}
			\annotatedFigureText{0.90,0.95}{black}{0.00}{\small{(B)}}
		\end{annotatedFigure}
		\begin{annotatedFigure}{\includegraphics[width=0.32\linewidth]{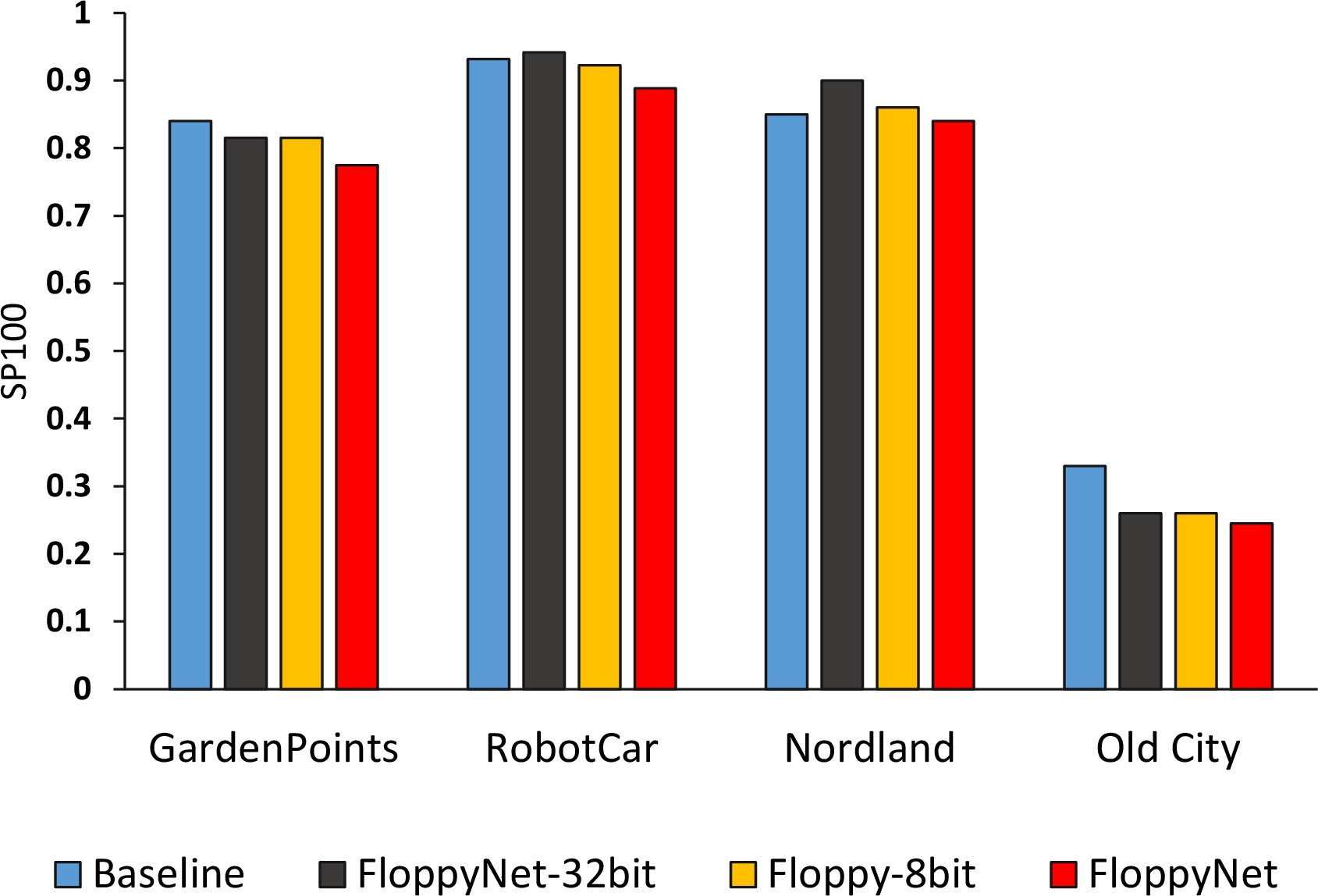}}
			\annotatedFigureText{0.90,0.95}{black}{0.00}{\small{(C)}}
		\end{annotatedFigure}	
\caption{FloppyNet's is compared on a variety of imaging conditions against the Baseline, BinaryNet and ShallowNet (A), full-precision networks (B) and several FloppyNet versions with different quantization levels (C).}
\label{fig:sp100}
\hspace{4ex} 
\end{figure*}

\begin{figure*}[!ht]
	\vspace*{1ex}
	\centering
		\begin{annotatedFigure}{\includegraphics[width=0.30\linewidth,height=4.1cm]{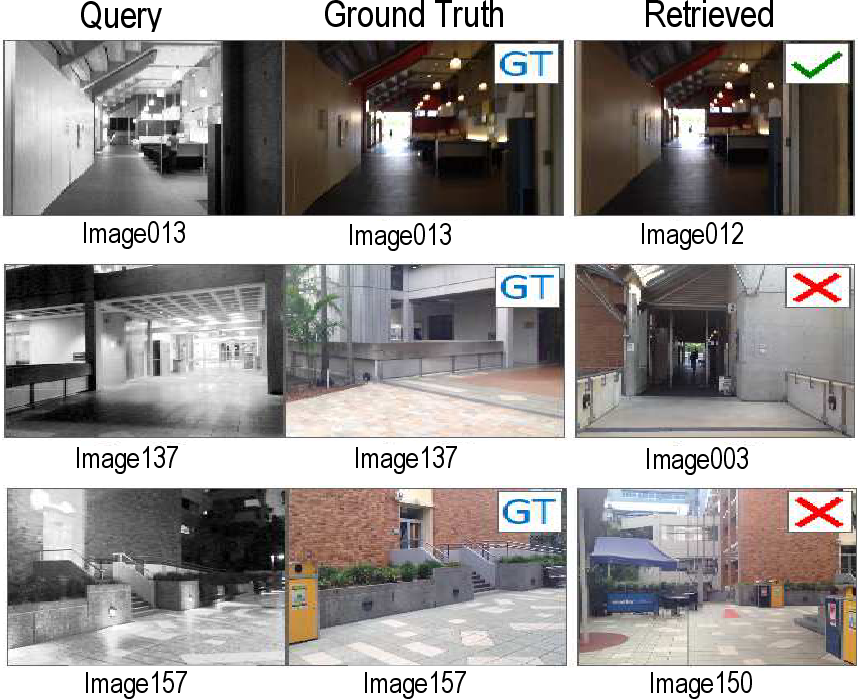}}
			\annotatedFigureText{0.95,1.05}{black}{0.00}{\small{(A)}}
			\annotatedFigureText{0.30,1.05}{black}{0.00}{\small{GardenPoints}}
		\end{annotatedFigure}
		\hspace{4ex}
		\begin{annotatedFigure}{\includegraphics[width=0.30\linewidth,height=4.1cm]{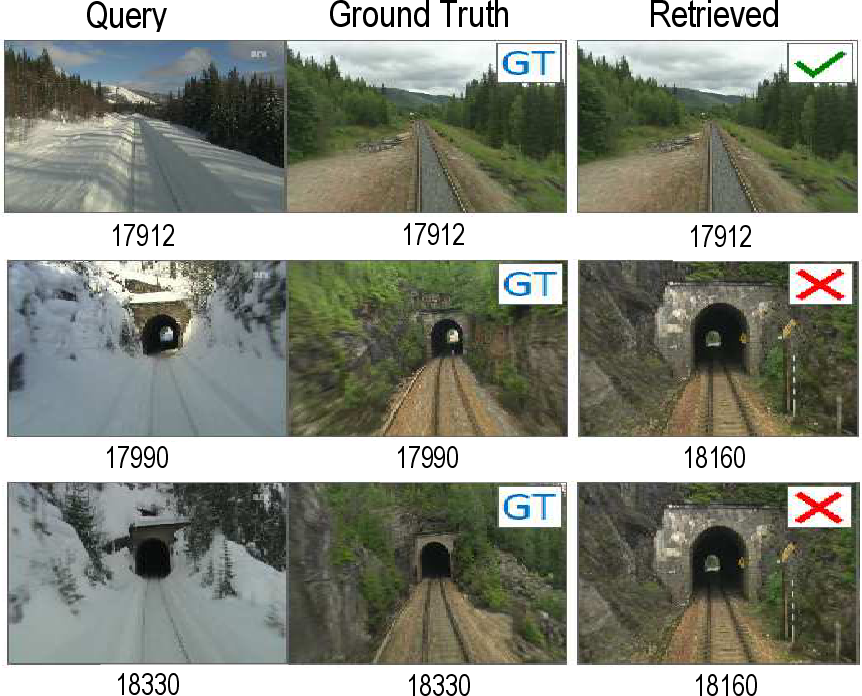}}
			\annotatedFigureText{0.95,1.05}{black}{0.00}{\small{(B)}}
			\annotatedFigureText{0.34,1.05}{black}{0.00}{\small{Nordland}}
		\end{annotatedFigure}
		\hspace{4ex}
		\begin{annotatedFigure}{\includegraphics[width=0.30\linewidth,height=4.1cm]{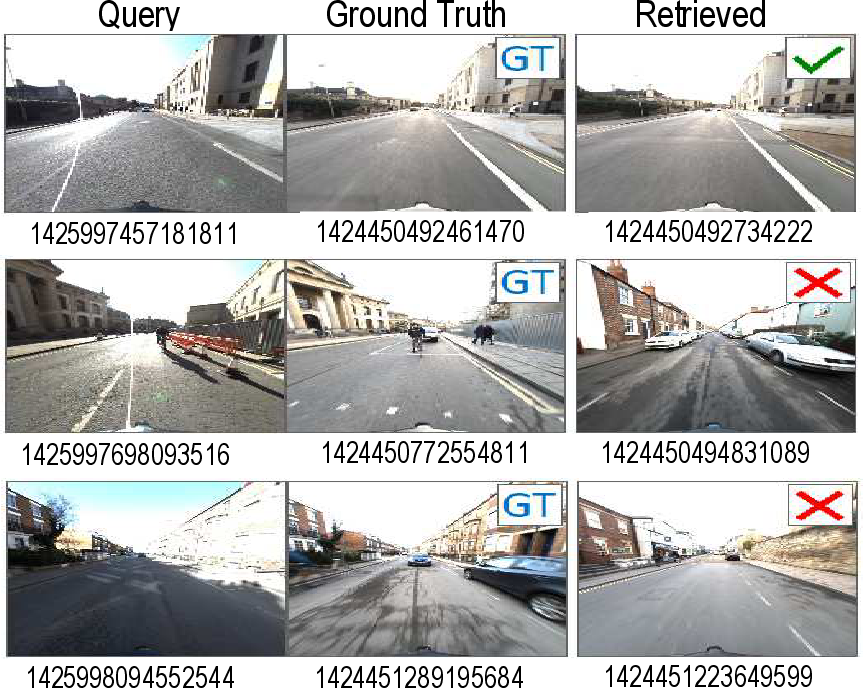}}
			\annotatedFigureText{0.91,1.05}{black}{0.00}{\small{(C)}}
			\annotatedFigureText{0.34,1.05}{black}{0.00}{\small{RobotCar}}
		\end{annotatedFigure}	
\caption{Some query results from GardenPoints (A), Nordland (B), and RobotCar Cross-Seasons (C) datasets.} 
\label{fig:scenes}
\hspace{4ex} 
\end{figure*}
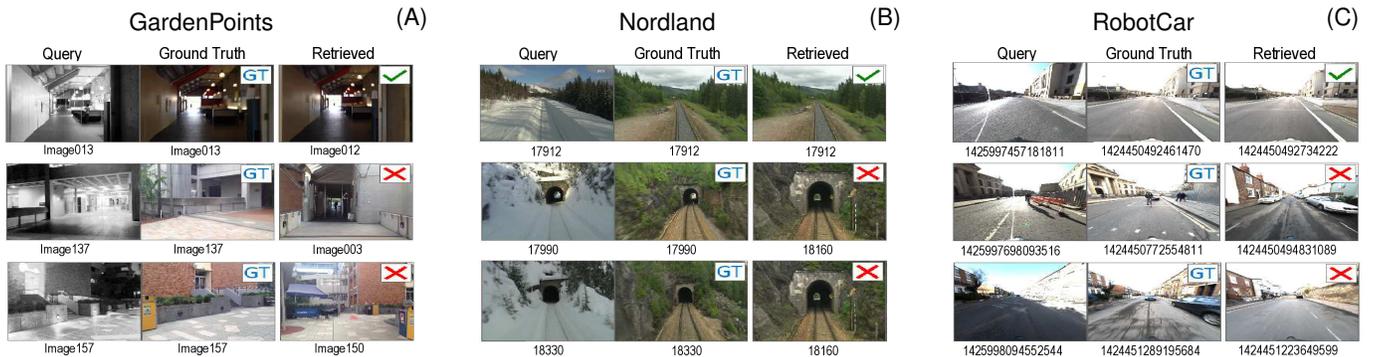

\subsection{FloppyNet}
\label{ssec:floppynet}

FloppyNet consists of three binary convolutional blocks and three pool layers, as shown in Fig. \ref{fig:floppynet}.
Binarization, jointly with depth reduction applied to Baseline, resulted in a compact model of $154$ KiB. The layers removed in the depth reduction step are \textit{conv3} and \textit{conv4}.
The output layer of FloppyNet is denoted as \textit{pool5} by convention. We kept the same name as in Baseline network since they have the same structure and both provide feature vectors with the same shape and element number: $6\times{}6\times{}256 = 9216$ elements. 

The primary motivation for FloppyNet is to reduce the model size further and shorten the inference latency. With two fewer convolutional layers, FloppyNet uses $33\%$ of the memory of BinaryNet at the \textit{pool5} layer, computing $39\%$ fewer MACs (Table \ref{tab:memory_floppy}).

Binarization and depth reduction cause a performance loss that is mostly compensated by tuning the FC stage properly for the training. Our best model is obtained with $256$ full-precision neurons in both \textit{fc6} and \textit{fc7}. This training approach's effectiveness is demonstrated in Section \ref{ssec:comparison} where FloppyNet is compared against ShallowNet, which is trained without tuning the FC stage.

\section{BNN Layers Analysis}
\label{ssec:layers}

The first question to answer when a convolutional network is employed as a feature extractor is: which layer is the most suitable to build a distinctive image descriptor?  
This section provides a VPR performance assessment of the features from every layer in both Baseline network (Section \ref{ssec:networks}) and its binary counterpart, BinaryNet. The results obtained drove the design of FloppyNet towards the use of \textit{pool5} as an output layer.

CNNs can learn features at different levels of abstraction.
Convolutional features retain some spatial information. However, as the depth increases, pool layers induce the loss of such a spatial information in favor of translation invariance.
In fully-connected layers, the activation of a neuron depends on every neuron in the previous level. 
Hence, the spatial information vanishes while improving the invariance to viewpoint changes and translation in particular \cite{7279659}. 
The second question we need to answer is how binarization impacts layers’ features and VPR matching performance.

The answers to these questions are given in Table \ref{tab:baseline_features} and Table \ref{tab:binarynet_features}, which show $S_{P100}$ for every layer of Baseline and BinaryNet, repectively. 
In Baseline, fully-connected layers and deeper convolutions handle viewpoint changes better than initial layers. 
\textit{Fc6} and \textit{fc7} obtain the highest performance under extreme viewpoint changes that characterize Old City. \textit{Pool5} is the best on GardenPoints, which includes mild viewpoint shifts other than day-light variations. 
On the other hand, shallower layers deal better with appearance changes. 
Nordland includes only seasonal variations, and the best layer is \textit{conv4} with $\ratio{} = 95\%$.
These results partially confirm findings of a previous study on a standard AlexNet \cite{7279659}, which indicates \textit{conv3} as the best layer to deal with appearance changes while \textit{pool5} and, in some cases, \textit{fc6} as the best choice to deal with viewpoint changes.

Binarization negatively affects VPR performance, but the characteristics of the layers are more or less unchanged. As shown in Table \ref{tab:binarynet_features}, \textit{pool5} achieves the highest performance on the same dataset as for Baseline. Similarly, fully-connected layers outperform the others on Old city.

Overall, \textit{pool5} is the layer that guarantees the highest average performance across the four datasets. 
The average $S_{P100}$ is $73.8\%$ for Baseline and $64.3\%$ for its binary counterpart. The gap is moderate ($9.5\%$), especially considering that BinaryNet at \textit{pool5} requires only $3.18\%$ of memory used by the baseline (Table \ref{tab:memory}). 
The average $S_{P100}$ is computed across all the datasets as they formed a single environment to simulate a workspace including a wide variety of viewpoint and appearance changes.
For a robot navigating such a workspace, $pool5$ features guarantee the most reliable and consistent VPR performance. Accordingly, we designed FloppyNet with the same shape progression of the features maps as in BinaryNet to have the output layer with similar characteristics as \textit{pool5}. As detailed in Section \ref{ssec:floppynet}, this is obtained by removing two inner convolutional blocks: \textit{conv3} and \textit{conv4}.

\section{VPR Performance Comparison}
\label{ssec:comparison}

FloppyNet is compared against several other networks. These include HybridNet, VGG-16, CALC, Baseline (Section \ref{ssec:networks}), BinaryNet, ShallowNet, and a 8-bit implementation of FloppyNet.
ShallowNet is the version of FloppyNet trained with regular fully-connected layers of 4096 binary neurons as described in Section \ref{ssec:depth_reduction}.
HybridNet \cite{chen2017deep} is a version of AlexNet with an additional convolutional block trained on ImageNet and tuned on SPED dataset \cite{chen2017deep}. To avoid training data influencing the results, we trained an HybridNet model by replacing ImageNet with Places365, which is the dataset used to train the other models considered for the experiments. The results for HybridNet are obtained with \textit{pool6} features. 
VGG-16 \cite{Simonyan14c} is a very deep network if compared with FloppyNet since it includes $13$ convolutional blocks. 
It is relevant to include VGG-16 in the comparison because several multi-staged VPR methods widely use it as a feature extractor. Some examples are R-MAC \cite{tolias2016rmac}, Cross-Region-Bow \cite{chen2017only} and NetVLAD \cite{arandjelovic2016netvlad}. 
The VGG-16 model has been trained from scratch using Places365, and the features used for the tests are from the very last pool layer. CALC \cite{Merrill2018RSS} is a lightweight CNN designed to address VPR with low resource requirements.
CALC includes about $137\,\text{K}$ parameters: a small fraction of $3.75\,\text{M}$ of Baseline and $1.24\,\text{M}$ of FloppyNet (Table VI). 
We used the model trained on Place365 shared by the authors. The 8-bit version of FloppyNet is included in the comparison to show that BNNs also scale well to 8-bits quantization, demonstrating potential applicability of binarization for VPR as a more efficient yet effective approach.

\subsection{Comparison with the Baseline}
\label{ssec:vsBaseline}
FloppyNet aims to achieve similar performance as the starting Baseline network (Section \ref{ssec:networks}) with higher efficiency. Fig. \ref{fig:sp100}.A shows comparative results between FloppyNet, Baseline and the intermediate design steps: BinaryNet and ShallowNet (Section \ref{ssec:floppynet}).
Binarization and depth reduction negatively impact VPR performance. BinaryNet and ShallowNet score the lowest $\ratio{}$ on every dataset, exhibiting a substantial gap from Baseline.
FloppyNet outperforms BinaryNet and ShallowNet on every dataset confirming that
 tuning  the fully-connected stage during training (Section \ref{ssec:fc_tuning}) mitigates the performance loss due to binarization and depth reduction.

In general Baseline has better performance than FloppyNet. However, on GardenPoints, Nordland, and Robotocar the difference is small and is due to a few places with some particular characteristics. 

The most difficult locations to recognize for FloppyNet are those presenting strong viewpoint variations. On Old City the $\ratio{}$ difference is $8.5\%$, which is the highest among the test datasets. An analysis on GardenPoints mismatches confirms such a FloppyNet's weakness. The $\ratio{}$ gap on GardenPoints is due mainly to lateral shift, which is generally mild but in a few locations, where FloppyNet fails while Baseline succeeds (Fig. \ref{fig:scenes}.A). 

FloppyNet scores almost the same $\ratio{}$ as Baseline on Nordland, $84\%$ vs $85\%$. 
The mismatches are mainly between locations showing tunnel entrances. Fig. \ref{fig:scenes}.B shows two examples of mismatch. FloppyNet retrieves the same reference image on two different queries including a tunnel.

FloppyNet scores a good $\ratio{}$ on RobotCar as well. The $\ratio{}$ difference with Baseline is caused by a series of wrong matches in the two locations shown in Fig. \ref{fig:scenes}.C. The high illumination contrast and the occlusions due to dynamic elements, such as cars and shadows, are possibly the cause of FloppyNet's failures. 

\subsection{Comparison with Full-Precision CNNs}
\label{ssec:floppy_vs_cnn}

Full-precision and deeper networks obtain better VPR performance than the proposed binary network (Fig. \ref{fig:sp100}.B). Substantial gaps are exhibited on Garden Points and Old City by VGG-16 and HybridNet, respectively. On the other hand, those two networks are far larger than FloppyNet even without considering the weights' precision.
FloppyNet scores an average $\ratio{}$ of $68.7\%$ using $1.24\,\text{M}$ binary parameters and computing $653\,\text{M}$ MACs. The highest average $\ratio{}$ is achieved by VGG-16, $77.9\%$. However, VGG-16 includes $14\,\text{M}$ weights and computes $15.3\,\text{B}$ floating-point MACs resulting in two orders of magnitude longer inference latency (Table \ref{tab:efficiency}). The small number of parameters penalizes CALC, which our network outperforms by a large margin on every dataset. Moreover,  its low VPR performance is not compensated by a sufficient computational efficiency as CALC is about two times slower than FloppyNet, as detailed in Section \ref{ssec:computation_spedd}. 

\subsection{Weight Precision Impact}
\label{ssec:precision_impact}
Fig. \ref{fig:sp100}.C presents a comparison between our best FloppyNet model against its 8-bit and full-precision versions to examine the impact of weight quantization on place recognition capabilities.
Increasing the quantization bits has a positive effect on VPR performance. FloppyNet-8bit outperforms FloppyNet in every test scenario and nearly closes the gap with Baseline (Section \ref{ssec:networks}) except on Old-City. Indeed, the reduced depth affects VPR performance on significant viewpoint changes, as discussed above in Section \ref{ssec:vsBaseline}. Overall, FloppyNet has slightly lower performance than the higher precision models except on GardenPoints, where the gap from the 8-bit implementation is the largest, $6\%$.
On the other hand, 8-bit quantization yields eight times larger models and almost doubles the inference latency and energy consumption, as shown in Section \ref{sec:deployment}. The higher resource demand of 8-bit quantization might be unsuitable for extremely cheap hardware, where binarization can be used instead.
Finally, further increasing the precision of the weights does not significantly improve the VPR capabilities of the proposed network. As shown in Fig. \ref{fig:sp100}.C, 8-bit and 32-bit models score similar $\ratio{}$ on every dataset. 

\subsection{Memory Efficiency}
\label{teb:efficiency}

Table \ref{tab:efficiency} shows the memory efficiency for all the compared networks. 
$S_{P100}$ is the average score on the four datasets. 
Binary networks have extremely low $\meff{}$ values compared to any full-precision network. FloppyNet requires $2.24\,\text{KiB}$ per $S_{P100}$ point, while CALC, the most memory-efficient CNN, requires $13.26\,\text{KiB}$. ShallowNet has the same size as FloppyNet but lower $S_{P100}$, hence it uses memory less efficiently: $\meff{} = 2.45\,\text{KiB}$.
FloppyNet-8bit performs better on average than the 1-bit model by $2.7\%$ but requires about eight times the memory. The considerably higher $\meff{} = 16.99\,\text{KiB}$ reflects this trend indicating that the increase of the model size does not correspond to a proportional increase in VPR performance.

\begin{figure}[t!]
    \centering
    \includegraphics[width=8.0cm]{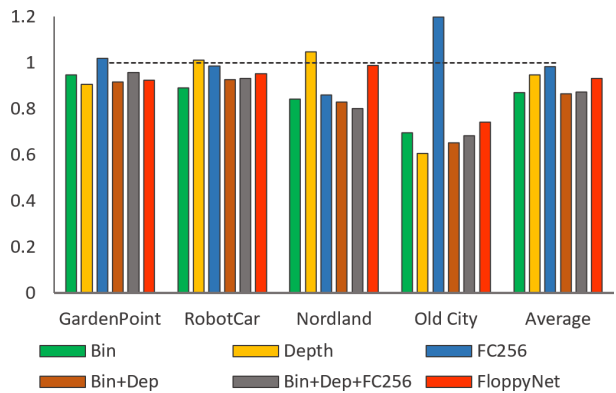}
    \caption{$S_{P100}$ is relative to Baseline (dotted line) of several combinations of the three techniques used by FloppyNet: depth reduction, binarization and training using a fully-connected stage with 256 neurons. The features used for VPR are from \textit{pool5} layer.}
    \label{fig:ablation}
\end{figure}

\begin{figure*}[h]
	\vspace*{1ex}
	\centering
	\textsc{\small\figtiTFlitefont{Inference Latency and Energy Usage}}\par\vspace*{2ex}
		\begin{annotatedFigure}{\includegraphics[width=0.43\linewidth]{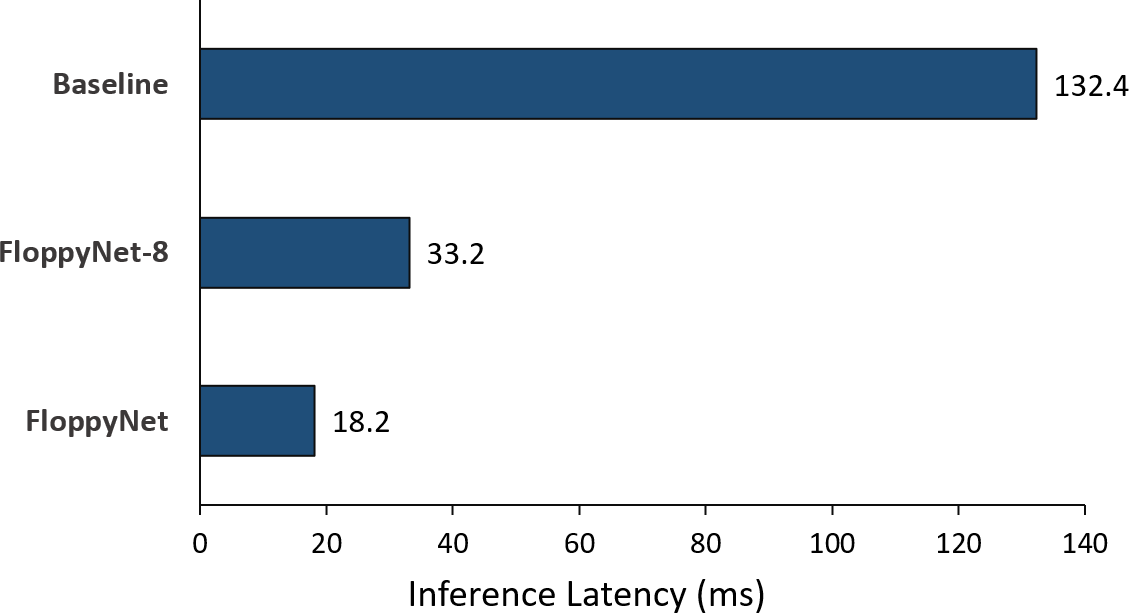}}
			\annotatedFigureText{0.05,0.00}{black}{0.00}{(A)}
		\end{annotatedFigure}
		\hspace{0.05\linewidth}
		\begin{annotatedFigure}{\includegraphics[width=0.43\linewidth]{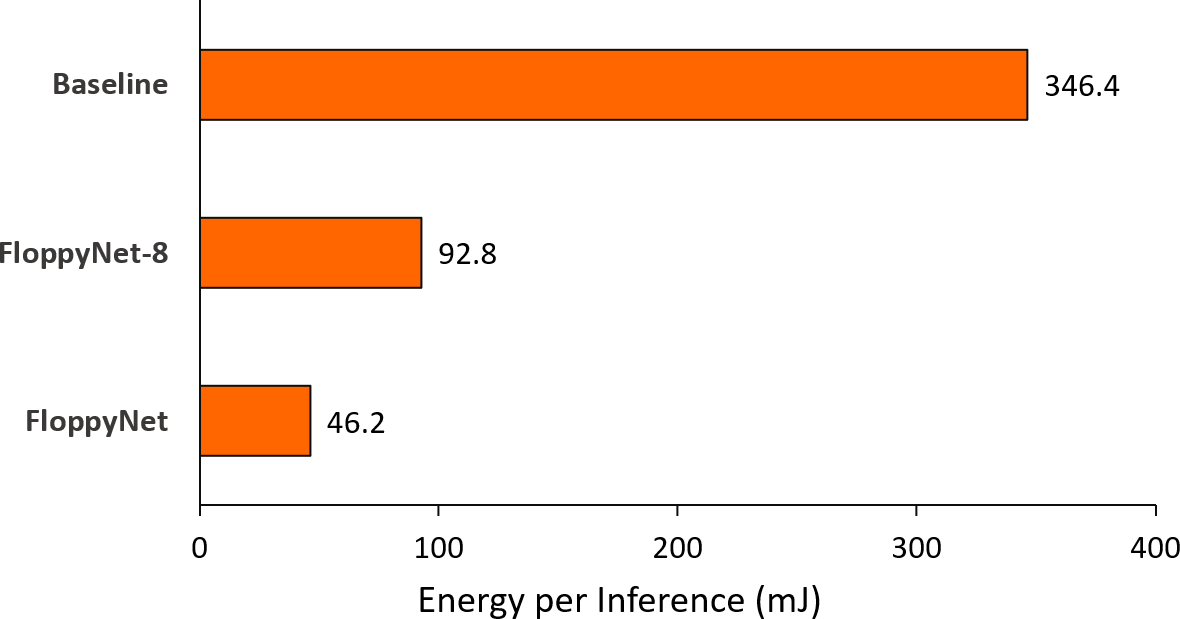}}
			\annotatedFigureText{0.05,0.00}{black}{0.00}{(B)}
		\end{annotatedFigure}
\caption{Inference latency (A) and energy usage (B) of FloppyNet models and Baseline executed with 4 threads on a Raspberry PI4.}
\label{fig:inference_latency}
\end{figure*}

\begin{table*}[!t]
  \centering 
  \caption{Models' performance and efficiency for Raspberry Pi 4 implementations.}
    \begin{tabular}{|l|c|c|ccc|ccc|cc|}
    \multicolumn{1}{r}{} & \multicolumn{1}{c}{\textbf{bits}} & \multicolumn{1}{c}{\textbf{av. SP100}} & \textbf{Params [M]} & \textbf{Size [KiB]} & \multicolumn{1}{c}{\textbf{$\boldsymbol{\meff{}}$ [KiB]}} & \textbf{MACs} & \textbf{$\boldsymbol{\tf{}}$ [ms]} & \multicolumn{1}{c}{\textbf{FPS}} & \textbf{$\boldsymbol{\power{}}$ [W]} & \multicolumn{1}{c}{\textbf{$\boldsymbol{\energy{}}$ [mJ]}} \\
    \hline
    \hline
    HybridNet & 32    & 75.4  & 5.07  & 16957 & 224.89 & 1098 M & 137.8 & 7.26  & 2.616 & 360.5 \bigstrut[t]\\
    Baseline & 32    & 73.8  & 3.75  & 14648 & 198.48 & 1077 M & 132.4 & 7.55  & 2.616 & 346.4 \\
    VGG-16 & 32    & 77.9  & 14.7  & 57487 & 737.96 & 15.3 B & 2211  & 0.45  & 2.616 & 5784 \\
    BinaryNet & 1     & 64.3  & 3.75  & 466   & 7.25  & 1077 M & 21.6  & 46.3  & 2.536 & 54.8 \\
    ShallowNet & 1     & 62.9  & 1.24  & 154   & 2.45  & 653 M & 18.2  & 54.95 & 2.536 & 46.2 \\
    FloppyNet & 1     & 68.7  & 1.24  & 154   & 2.24  & 653 M & 18.2  & 54.95 & 2.536 & 46.2 \\
    FloppyNet-8 & 8     & 71.4  & 1.24  & 1213  & 16.99 & 653 M & 33.2  & 30.12 & 2.876 & 95.5 \\
    FloppyNet-32 & 32    & 72.9  & 1.24  & 4843  & 66.43 & 653 M & 76.1  & 13.14 & 2.616 & 199.1 \\
    CALC  & 32    & 40.5  & 0.137 & 537   & 13.26 & 186 M & 45.1  & 22.17 & 2.616 & 118.0 \\
    \hline
    \end{tabular}%
  \label{tab:efficiency}%
\end{table*}%

\subsection{Binarization, Depth Reduction and FC-256}
\label{ssec:ablation}
Fig. \ref{fig:ablation} shows $S_{P100}$ relatively to Baseline (Section \ref{ssec:networks}) resulting from using binarization (Bin), depth reduction (Depth), and FC stage tuning (FC256) separately and their relevant combinations. The features used to obtain the results are from the \textit{pool5} layer.

Depth reduction (Depth) yields a full-precision network with better performance on Nordland and RobotCar.
Depth reduction makes the output layer of a model retain more spatial information compared to Baseline. Shallower layers are more suitable to deal with appearance changes (Table \ref{tab:baseline_features}). Hence, the better performance on Nordland and Robotcar datasets that present significant appearance changes while none or mild viewpoint variations. Training a full-precision model with 256 neurons in the FC stage (FC256) helps VPR significantly to tackle extreme 6-DOF viewpoint variations. FC256 model achieves $25\%$ higher $S_{P100}$ on Old City than the original model trained with 4096 neurons in the FC stage. In classifiers, the FC stage is usually sized as large as possible to maximize accuracy while avoiding overfitting. Conversely, the empirical evidence shows that VPR benefits from a smaller FC stage. These results suggest that the FC stage has different roles in classification and VPR tasks.

Depth reduction does not help $S_{p100}$ scores of the binarized models. The values of $S_{p100}$ for binary (bin) and shallow binary networks (bin+dep) are very close to each other on every test dataset (Nordland in particular), which is rather unexpected considering the full-precision case (Depth).
The red bars in Fig. \ref{fig:ablation} represent FloppyNet, which implements all the steps of the proposed approach. 
The addition of FC stage tuning counters the $S_{p100}$ loss due to binarization and depth reduction in every tested scenario (except Garden Point) supporting the effectiveness of the proposed training approach. 


\begin{figure*}[th]
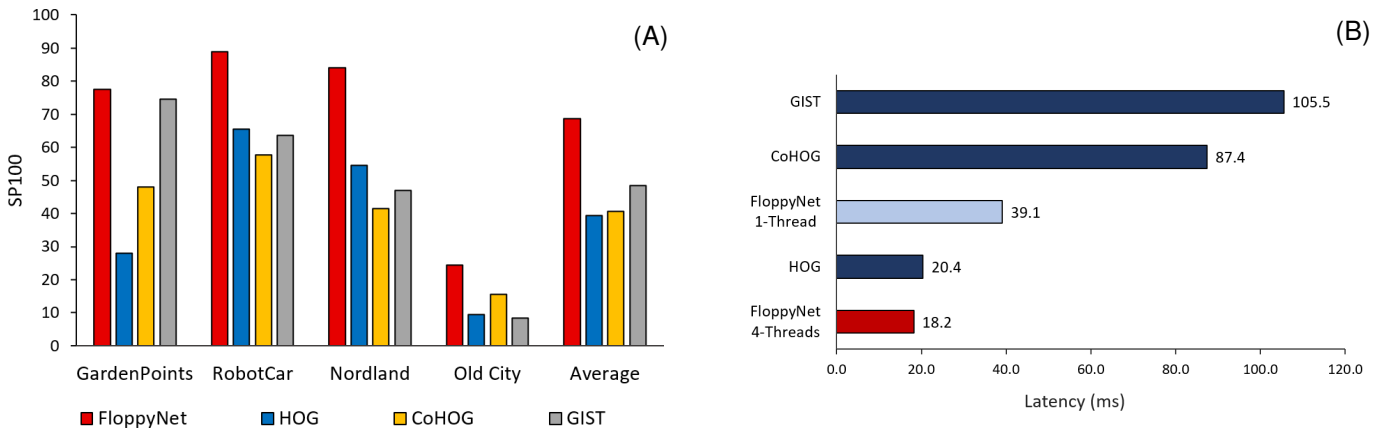

	\vspace*{1ex}
	\centering
		\begin{annotatedFigure}{\includegraphics[width=0.48\linewidth]{\fullpath{SP100_D_mine}}}
			\annotatedFigureText{0.95,0.90}{black}{0.00}{(A)}
		\end{annotatedFigure}
		\hspace{0.05\linewidth}
		\begin{annotatedFigure}{\includegraphics[width=0.45\linewidth]{\fullpath{inference_latency_hand_mine}}}
			\annotatedFigureText{0.95,0.95}{black}{0.00}{(B)}
		\end{annotatedFigure}
\caption{VPR performance and computational latency of HOG, CoHOG and GIST compared with FloppyNet. The latency is measured on a Rasperry PI4.}
\label{fig:hand}
\end{figure*}

\section{Benchmark Analysis}
\label{sec:deployment}
The software framework used to deploy binary models is Larq Compute Engine kernel (LCE) \cite{bannink2020larq}. LCE consists of a model compiler and a kernel to compute binary convolutions within the Tensorflow Lite runtime environment (TFlite). \cite{tfliteweb,david2020tensorflow}
LCE was a natural choice as it is part of the Larq  ecosystem \cite{larq} we used to train our models.
The 8-bit and full-precision implementations use the built-in TFlite compute kernel.

The evaluation criteria are the inference latency and power usage. 
Inference latency indicates the time required to complete an inference, namely to compute an image representation. Power usage is measured on a model execution and is used to determine the energy cost per inference (Eq. \ref{eq:energy}).
The platform employed for the experiments is a Raspberry PI4 (RPI4) that sits on an ARMv8 Cortex-A72 running at $1.5$ GHz \cite{rpi4-spec}. The operating system is Ubuntu 20.04 Linux, 64-bit.

\subsection{Inference Latency and Computation Speed-Up}
\label{ssec:computation_spedd}

Fig. \ref{fig:inference_latency}.A compares the inference latency of Baseline (Section \ref{ssec:networks}) and two FloppyNet implementations: 1-bit and 8-bit. 
The values reported in the figure are an average of 100 runs using four threads, namely employing all the cores available in the RPI4's CPU. 
FloppyNet has a latency of $18.2\,ms$, which is about seven times faster than Baseline ($132.4\,ms$). 

Table \ref{tab:efficiency} shows the inference latency for all the tested models. FloppyNet is considerably faster than any other full precision CNN including CALC, which computes only $186\,\text{M}$ MACs, $28.5\%$ of FloppyNet's MACs. However, it takes a considerably longer inference time than our network, $45.1\,ms$ against $18.2\,ms$, while achieving considerably lower VPR performance (Fig. \ref{fig:sp100}.B). BinaryNet latency enables an analysis of the speed-up contributions provided by binarization and depth reduction. BinaryNet has the same structure as Baseline except for the binary weights and runs in $21.6\,ms$, resulting in six times faster than its full-precision counterpart, Baseline. Depth reduction removes two convolutions from BinaryNet, shortening the execution by a further $16\%$.
Latency measures demonstrate that FloppyNet also scales well to 8-bit quantization. FloppyNet-8bit complete an inference in $33.2\,ms$, resulting almost twice slower than FloppyNet. 
%

\subsection{Power Usage}
\label{ssec:power}
The power usage is measured with an ampere meter connected to the USB-C power port of the RPI4.
The energy spent per inference by a model, $\energy{}$, depends on the power absorbed by the RPI4, $\power$, and the inference latency, $\tf{}$ (Eq. \ref{eq:energy}).
$\power$ is stable during an inference depending only on a model's weights precision because of the computational kernel used: LCE for binary models and TFlite for 8-bit and full-precision models. 
The measured $\power{}$ is $2.53\,W$, $2.62\,W$ and $2.88\,W$ for binary, 32-bit and 8-bit models, respectively.
As $\tf{}$ varies in a wider range than $\power$ (Fig. \ref{fig:inference_latency}), $\energy{}$ depends mainly on a model's inference latency.
The rightmost columns of Table \ref{tab:efficiency} show the power absorption and inference energy for every model. BNNs consumes less energy than any other considered network. In particular, 
FloppyNet uses $46.1\,mJ$ per inference, which is considerably more energy-efficient than the 8-bit implementation and Baseline spend of $92\,mJ$ and $346\,mJ$, respectively.

\section{Comparison with Handcrafted Descriptors}
\label{sec:handcrafted}

This section compares FloppyNet with several handcrafted image descriptors relevant for VPR applications \cite{zaffar2020vpr}.
They include HOG \cite{dalal2005histograms}, GIST \cite{oliva2006building} and CoHOG \cite{zaffar2020cohog}. The results show that our model has significantly better VPR capabilities while having comparable or higher computational efficiency than the considered handcrafted descriptors.

\subsection{Handcrafted Descriptors Setup}

The platform employed for measuring the processing time is a Raspberry PI4 (RPI4), as in the previous section. The handcrafted descriptors setting used for the experiments are as follows:

\subsubsection{HOG}

\noindent we used the \text{OpenCV 4.5.0} implementation with a cell size of $16 \times 16$ and block size of $32 \times 32$, as suggested in \cite{zaffar2020vpr}. The input image size is set to $256 \times 256$ pixels, which is similar to FloppyNet's input size of $277 \times 277$ pixels.

\subsubsection{CoHOG}

\noindent we used the code shared by the authors with their recommended settings \cite{chog-repo}: cell size of $16 \times 16$, $8$ bins and entropy threshold of $0.4$. The image size is $256 \times 256$ pixels. 

\subsubsection{Gist}

it is available in the \textit{C} library \textit{Lear's GIST} through the \textit{pyleargist} python wrapper \cite{pyleargist}. 
Gist is used with the parameters indicated by the authors \cite{gist-torralba}: $4$ blocks and $8$ orientations per scale. We set the image size to $128 \times 128$ pixels to keep the Gist's latency comparable with the other methods. Indeed, using $256 \times 256$ images extends Gist's processing time by roughly four times.

\subsection{Results Discussion}

Fig. \ref{fig:hand} shows the results. FloppyNet achieves substantially better VPR performance on every dataset scoring an average $\ratio{}$ of $68.7\%$. Gist achieves the highest VPR performance among the handcrafted descriptors but results in the slowest one taking $105\,ms$ to process an image.

Our network is the fastest technique when running using all the four RPI4's cores. However, Gist and HOG implementations cannot run on multiple threads. We reported the inference latency for FloppyNet running on a single thread in Fig. \ref{fig:hand}.B for a fair comparison. Only HOG runs faster than our network taking $20.4\,ms$ to process an image  instead of the $39.1\,ms$ required by FloppyNet (1-thread). On the other hand, HOG exhibits a wide VPR performance gap on every dataset. HOG scores an average $\ratio{}$ of $39.4\%$, whereas FloppyNet achieves $\ratio{} = 68.7\%$. We conclude that the shorter latency of HOG does not compensate for the poor VPR performance it achives compared to our network.

\section{Conclusions and Future Works}
\label{sec:conclusions}

This paper proposes FloppyNet, a compact binary network, to solve the VPR problem. FloppyNet achieves comparable VPR performance to deeper and full-precision networks in changing environments with drastically lower memory requirements and substantial computational speed-up.
Such lightweight networks open up several opportunities for embedded systems and edge computing in general. FloppyNet may be employed to enable VPR on very cheap hardware or replace standard CNNs to free up resources to allocate for additional functionalities to improve a robot's navigation system or increase the frame rate on low-cost embedded applications.
For example, NetVLAD is a two-stage image descriptor that uses VGG  to extract image features that are subsequently post-processed to compute a robust image representation. VGG is a large network that requires a relatively long time to extract image features. If a BNN such as FloppyNet is used instead, NetVLAD's memory requirements and computational efficiency improve dramatically. 
This example suggests that a natural extension of this work investigates the applicability of BNNs in multi-stage descriptors that use a CNN as a feature extractor.

\bibliographystyle{abbrv}
\bibliography{bib}

\begin{thebibliography}{10}

\bibitem{arm_thoughput}
{Arm Cortex-A76 Software Optimization Guide}.
\newblock \url{https://developer.arm.com/documentation/swog307215/a}.
\newblock Accessed: 2021-03-03.

\bibitem{chog-repo}
{CoHOG source code}.
\newblock
  \url{https://github.com/MubarizZaffar/VPR-Bench/tree/main/VPR_Techniques/CoHOG_Python}.
\newblock Accessed: 2021-11-06.

\bibitem{cuda_thoughput}
{CUDA Toolkit Documentation}.
\newblock
  \url{https://docs.nvidia.com/cuda/cuda-c-programming-guide/index.html#arithmetic-instructions}.
\newblock Accessed: 2020-07-26.

\bibitem{gist-torralba}
{GIST MATLAB implementation}.
\newblock \url{http://people.csail.mit.edu/torralba/code/spatialenvelope/}.
\newblock Accessed: 2021-11-06.

\bibitem{places365-gith}
{Places365 DevKit}.
\newblock \url{https://github.com/zhoubolei/places_devkit}.
\newblock Accessed: 2020-07-26.

\bibitem{pyleargist}
{pyleargist}.
\newblock \url{https://pypi.org/project/pyleargist/}.
\newblock Accessed: 2021-11-06.

\bibitem{pynq.io}
{PYNQ}.
\newblock \url{http://www.pynq.io/}.
\newblock Accessed: 2021-04-26.

\bibitem{rpi4-spec}
{Raspberry Pi 4 Tech Specs}.
\newblock
  \url{https://www.raspberrypi.org/products/raspberry-pi-4-model-b/specifications/}.
\newblock Accessed: 2021-03-06.

\bibitem{tfliteweb}
{Tensorflow Lite}.
\newblock \url{https://www.tensorflow.org/lite}.
\newblock Accessed: 2021-03-05.

\bibitem{V4RLdata}
{V4RL Wide-baseline Place Recognition Dataset}.
\newblock
  \url{https://github.com/VIS4ROB-lab/place_recognition_dataset_ral2019}.
\newblock Accessed: 2019-04-04.

\bibitem{tensorflow2015-whitepaper}
M.~Abadi, A.~Agarwal, P.~Barham, E.~Brevdo, Z.~Chen, C.~Citro, G.~S. Corrado,
  A.~Davis, J.~Dean, M.~Devin, S.~Ghemawat, I.~Goodfellow, A.~Harp, G.~Irving,
  M.~Isard, Y.~Jia, R.~Jozefowicz, L.~Kaiser, M.~Kudlur, J.~Levenberg,
  D.~Man\'{e}, R.~Monga, S.~Moore, D.~Murray, C.~Olah, M.~Schuster, J.~Shlens,
  B.~Steiner, I.~Sutskever, K.~Talwar, P.~Tucker, V.~Vanhoucke, V.~Vasudevan,
  F.~Vi\'{e}gas, O.~Vinyals, P.~Warden, M.~Wattenberg, M.~Wicke, Y.~Yu, and
  X.~Zheng.
\newblock {TensorFlow}: Large-scale machine learning on heterogeneous systems,
  2015.
\newblock Software available from tensorflow.org.

\bibitem{alizadeh2018a}
M.~Alizadeh, J.~Fernández-Marqués, N.~D. Lane, and Y.~Gal.
\newblock An empirical study of binary neural networks' optimisation.
\newblock In {\em International Conference on Learning Representations}, 2019.

\bibitem{arandjelovic2016netvlad}
R.~Arandjelovic, P.~Gronat, A.~Torii, T.~Pajdla, and J.~Sivic.
\newblock Netvlad: {CNN} architecture for weakly supervised place recognition.
\newblock In {\em Proceedings of the IEEE Conference on Computer Vision and
  Pattern Recognition}, pages 5297--5307, 2016.

\bibitem{bai2018sequence}
D.~Bai, C.~Wang, B.~Zhang, X.~Yi, and X.~Yang.
\newblock Sequence searching with cnn features for robust and fast visual place
  recognition.
\newblock {\em Computers \& Graphics}, 70:270--280, 2018.

\bibitem{bannink2020larq}
T.~Bannink, A.~Hillier, L.~Geiger, T.~de~Bruin, L.~Overweel, J.~Neeven, and
  K.~Helwegen.
\newblock Larq compute engine: Design, benchmark, and deploy state-of-the-art
  binarized neural networks, 2020.

\bibitem{baya_speeded-up_2008}
H.~Bay, A.~Ess, T.~Tuytelaars, and L.~Van~Gool.
\newblock Speeded-up robust features ({SURF}).
\newblock {\em Computer Vision and Image Understanding}, 110(3):346--359, 2008.

\bibitem{bengio2013estimating}
Y.~Bengio, N.~L{\'{e}}onard, and A.~C. Courville.
\newblock Estimating or propagating gradients through stochastic neurons for
  conditional computation.
\newblock {\em CoRR}, abs/1308.3432, 2013.

\bibitem{bethge2019simplicity}
J.~Bethge, H.~Yang, M.~Bornstein, and C.~Meinel.
\newblock Back to simplicity: How to train accurate bnns from scratch?
\newblock {\em CoRR}, abs/1906.08637, 2019.

\bibitem{8506339}
S.~Bianco, R.~Cadene, L.~Celona, and P.~Napoletano.
\newblock Benchmark analysis of representative deep neural network
  architectures.
\newblock {\em IEEE Access}, 6:64270--64277, 2018.

\bibitem{cai2019once}
H.~Cai, C.~Gan, T.~Wang, Z.~Zhang, and S.~Han.
\newblock Once for all: Train one network and specialize it for efficient
  deployment.
\newblock In {\em International Conference on Learning Representations}, 2020.

\bibitem{chen2015mxnet}
T.~Chen, M.~Li, Y.~Li, M.~Lin, N.~Wang, M.~Wang, T.~Xiao, B.~Xu, C.~Zhang, and
  Z.~Zhang.
\newblock Mxnet: {A} flexible and efficient machine learning library for
  heterogeneous distributed systems.
\newblock {\em CoRR}, abs/1512.01274, 2015.

\bibitem{chen2018tvm}
T.~Chen, T.~Moreau, Z.~Jiang, L.~Zheng, E.~Yan, H.~Shen, M.~Cowan, L.~Wang,
  Y.~Hu, L.~Ceze, et~al.
\newblock $\{$TVM$\}$: An automated end-to-end optimizing compiler for deep
  learning.
\newblock In {\em 13th $\{$USENIX$\}$ Symposium on Operating Systems Design and
  Implementation ($\{$OSDI$\}$ 18)}, pages 578--594, 2018.

\bibitem{chen2017deep}
Z.~Chen, A.~Jacobson, N.~S{\"u}nderhauf, B.~Upcroft, L.~Liu, C.~Shen, I.~Reid,
  and M.~Milford.
\newblock Deep learning features at scale for visual place recognition.
\newblock In {\em 2017 IEEE International Conference on Robotics and Automation
  (ICRA)}, pages 3223--3230. IEEE, 2017.

\bibitem{chen2017only}
Z.~Chen, F.~Maffra, I.~Sa, and M.~Chli.
\newblock Only look once, mining distinctive landmarks from convnet for visual
  place recognition.
\newblock In {\em 2017 IEEE/RSJ International Conference on Intelligent Robots
  and Systems (IROS)}, pages 9--16. IEEE, 2017.

\bibitem{chollet2015keras}
F.~Chollet et~al.
\newblock Keras.
\newblock \url{https://github.com/fchollet/keras}, 2015.

\bibitem{courbariaux2016binarized}
M.~Courbariaux and Y.~Bengio.
\newblock Binarynet: Training deep neural networks with weights and activations
  constrained to +1 or -1.
\newblock {\em CoRR}, abs/1602.02830, 2016.

\bibitem{courbariaux2014training}
M.~Courbariaux, Y.~Bengio, and J.-P. David.
\newblock Training deep neural networks with low precision multiplications.
\newblock {\em arXiv e-prints}, pages arXiv--1412, 2014.

\bibitem{cummins2011appearance}
M.~Cummins and P.~Newman.
\newblock Appearance-only slam at large scale with fab-map 2.0.
\newblock {\em The International Journal of Robotics Research},
  30(9):1100--1123, 2011.

\bibitem{dalal2005histograms}
N.~Dalal and B.~Triggs.
\newblock Histograms of oriented gradients for human detection.
\newblock In {\em 2005 IEEE Computer Society Conference on Computer Vision and
  Pattern Recognition (CVPR'05)}, volume~1, pages 886--893 vol. 1, 2005.

\bibitem{david2020tensorflow}
R.~David, J.~Duke, A.~Jain, V.~J. Reddi, N.~Jeffries, J.~Li, N.~Kreeger,
  I.~Nappier, M.~Natraj, S.~Regev, R.~Rhodes, T.~Wang, and P.~Warden.
\newblock Tensorflow lite micro: Embedded machine learning on tinyml systems.
\newblock {\em CoRR}, abs/2010.08678, 2020.

\bibitem{esser2019learned}
S.~K. Esser, J.~L. McKinstry, D.~Bablani, R.~Appuswamy, and D.~S. Modha.
\newblock Learned step size quantization.
\newblock {\em CoRR}, abs/1902.08153, 2019.

\bibitem{ferrarini2019visual}
B.~Ferrarini, M.~Waheed, S.~Waheed, S.~Ehsan, M.~Milford, and K.~D.
  McDonald-Maier.
\newblock Visual place recognition for aerial robotics: Exploring
  accuracy-computation trade-off for local image descriptors.
\newblock In {\em 2019 NASA/ESA Conference on Adaptive Hardware and Systems
  (AHS)}, pages 103--108. IEEE, 2019.

\bibitem{EP}
B.~{Ferrarini}, M.~{Waheed}, S.~{Waheed}, S.~{Ehsan}, M.~J. {Milford}, and
  K.~D. {McDonald-Maier}.
\newblock Exploring performance bounds of visual place recognition using
  extended precision.
\newblock {\em IEEE Robotics and Automation Letters}, 5(2):1688--1695, April
  2020.

\bibitem{freeman1995orientation}
W.~T. Freeman and M.~Roth.
\newblock Orientation histograms for hand gesture recognition.
\newblock In {\em International workshop on automatic face and gesture
  recognition}, volume~12, pages 296--301, 1995.

\bibitem{MLSYS2020_2a79ea27}
J.~Fromm, M.~Cowan, M.~Philipose, L.~Ceze, and S.~Patel.
\newblock Riptide: Fast end-to-end binarized neural networks.
\newblock In I.~Dhillon, D.~Papailiopoulos, and V.~Sze, editors, {\em
  Proceedings of Machine Learning and Systems}, volume~2, pages 379--389, 2020.

\bibitem{larq}
L.~Geiger and P.~Team.
\newblock Larq: An open-source library for training binarized neural networks.
\newblock {\em Journal of Open Source Software}, 5(45):1746, Jan. 2020.

\bibitem{han2015learning}
S.~Han, J.~Pool, J.~Tran, and W.~J. Dally.
\newblock Learning both weights and connections for efficient neural networks.
\newblock In {\em Proceedings of the 28th International Conference on Neural
  Information Processing Systems-Volume 1}, pages 1135--1143, 2015.

\bibitem{hassibi1992second}
B.~Hassibi and D.~G. Stork.
\newblock Second order derivatives for network pruning: optimal brain surgeon.
\newblock In {\em Proceedings of the 5th International Conference on Neural
  Information Processing Systems}, pages 164--171, 1992.

\bibitem{7279659}
Y.~{Hou}, H.~{Zhang}, and S.~{Zhou}.
\newblock Convolutional neural network-based image representation for visual
  loop closure detection.
\newblock In {\em 2015 IEEE International Conference on Information and
  Automation}, pages 2238--2245, 2015.

\bibitem{HowardZCKWWAA17}
A.~G. Howard, M.~Zhu, B.~Chen, D.~Kalenichenko, W.~Wang, T.~Weyand,
  M.~Andreetto, and H.~Adam.
\newblock Mobilenets: Efficient convolutional neural networks for mobile vision
  applications.
\newblock {\em CoRR}, abs/1704.04861, 2017.

\bibitem{hubara2017quantized}
I.~Hubara, M.~Courbariaux, D.~Soudry, R.~El-Yaniv, and Y.~Bengio.
\newblock Quantized neural networks: Training neural networks with low
  precision weights and activations.
\newblock {\em The Journal of Machine Learning Research}, 18(1):6869--6898,
  2017.

\bibitem{batchnorm}
S.~Ioffe and C.~Szegedy.
\newblock Batch normalization: Accelerating deep network training by reducing
  internal covariate shift.
\newblock In {\em International Conference on Machine Learning}, pages
  448--456, 2015.

\bibitem{jegou2010aggregating}
H.~J{\'e}gou, M.~Douze, C.~Schmid, and P.~P{\'e}rez.
\newblock Aggregating local descriptors into a compact image representation.
\newblock In {\em CVPR 2010-23rd IEEE Conference on Computer Vision \& Pattern
  Recognition}, pages 3304--3311. IEEE Computer Society, 2010.

\bibitem{8695662}
A.~Karki, C.~Palangotu~Keshava, S.~Mysore~Shivakumar, J.~Skow,
  G.~Madhukeshwar~Hegde, and H.~Jeon.
\newblock Tango: A deep neural network benchmark suite for various
  accelerators.
\newblock In {\em 2019 IEEE International Symposium on Performance Analysis of
  Systems and Software (ISPASS)}, pages 137--138, 2019.

\bibitem{khaliq2018holistic}
A.~{Khaliq}, S.~{Ehsan}, Z.~{Chen}, M.~{Milford}, and K.~{McDonald-Maier}.
\newblock A holistic visual place recognition approach using lightweight cnns
  for significant viewpoint and appearance changes.
\newblock {\em IEEE Transactions on Robotics}, pages 1--9, 2019.

\bibitem{khaliq2019camal}
A.~Khaliq, S.~Ehsan, M.~Milford, and K.~D. McDonald{-}Maier.
\newblock {CAMAL:} context-aware multi-scale attention framework for
  lightweight visual place recognition.
\newblock {\em CoRR}, abs/1909.08153, 2019.

\bibitem{khan2020survey}
A.~Khan, A.~Sohail, U.~Zahoora, and A.~S. Qureshi.
\newblock A survey of the recent architectures of deep convolutional neural
  networks.
\newblock {\em Artificial Intelligence Review}, pages 1--62, 2020.

\bibitem{krizhevsky2012imagenet}
A.~Krizhevsky, I.~Sutskever, and G.~E. Hinton.
\newblock Imagenet classification with deep {Convolutional Neural Networks}.
\newblock In {\em Advances in neural information processing systems}, pages
  1097--1105, 2012.

\bibitem{larsson2019cross}
M.~M. Larsson, E.~Stenborg, L.~Hammarstrand, M.~Pollefeys, T.~Sattler, and
  F.~Kahl.
\newblock A cross-season correspondence dataset for robust semantic
  segmentation.
\newblock In {\em 2019 IEEE/CVF Conference on Computer Vision and Pattern
  Recognition (CVPR)}, pages 9524--9534. IEEE, 2019.

\bibitem{lecun1990optimal}
Y.~LeCun, J.~S. Denker, and S.~A. Solla.
\newblock Optimal brain damage.
\newblock In {\em Advances in neural information processing systems}, pages
  598--605, 1990.

\bibitem{li2016ternary}
F.~Li and B.~Liu.
\newblock Ternary weight networks.
\newblock {\em CoRR}, abs/1605.04711, 2016.

\bibitem{lowe2004distinctive}
D.~G. Lowe.
\newblock Distinctive image features from scale-invariant keypoints.
\newblock {\em International journal of computer vision}, 60(2):91--110, 2004.

\bibitem{lowry2016supervised}
S.~Lowry and M.~J. Milford.
\newblock Supervised and unsupervised linear learning techniques for visual
  place recognition in changing environments.
\newblock {\em IEEE Transactions on Robotics}, 32(3):600--613, 2016.

\bibitem{RobotCarDatasetIJRR}
W.~Maddern, G.~Pascoe, C.~Linegar, and P.~Newman.
\newblock {1 Year, 1000km: The Oxford RobotCar Dataset}.
\newblock {\em The International Journal of Robotics Research (IJRR)},
  36(1):3--15, 2017.

\bibitem{maffra2018tolerant}
F.~Maffra, Z.~Chen, and M.~Chli.
\newblock Viewpoint-tolerant place recognition combining 2d and 3d information
  for uav navigation.
\newblock In {\em 2018 IEEE International Conference on Robotics and Automation
  (ICRA)}, pages 2542--2549. IEEE, 2018.

\bibitem{maffra2019real}
F.~Maffra, L.~Teixeira, Z.~Chen, and M.~Chli.
\newblock Real-time wide-baseline place recognition using depth completion.
\newblock {\em IEEE Robotics and Automation Letters}, 2019.

\bibitem{mcmanus2014scene}
C.~McManus, B.~Upcroft, and P.~Newmann.
\newblock Scene signatures: Localised and point-less features for localisation.
\newblock In {\em Proceedings of Robotics: Science and Systems}, Berkeley, USA,
  July 2014.

\bibitem{Merrill2018RSS}
N.~Merrill and G.~Huang.
\newblock Lightweight unsupervised deep loop closure.
\newblock In {\em Proceedings of Robotics: Science and Systems}, Pittsburgh,
  Pennsylvania, June 2018.

\bibitem{milford2012seqslam}
M.~J. Milford and G.~F. Wyeth.
\newblock Seqslam: Visual route-based navigation for sunny summer days and
  stormy winter nights.
\newblock In {\em 2012 IEEE International Conference on Robotics and
  Automation}, pages 1643--1649. IEEE, 2012.

\bibitem{murillo2007surf}
A.~C. Murillo, J.~J. Guerrero, and C.~Sagues.
\newblock Surf features for efficient robot localization with omnidirectional
  images.
\newblock In {\em Proceedings 2007 IEEE International Conference on Robotics
  and Automation}, pages 3901--3907. IEEE, 2007.

\bibitem{oliva2006building}
A.~Oliva and A.~Torralba.
\newblock Building the gist of a scene: The role of global image features in
  recognition.
\newblock {\em Progress in brain research}, 155:23--36, 2006.

\bibitem{8942095}
I.~Palit, Q.~Lou, R.~Perricone, M.~Niemier, and X.~S. Hu.
\newblock A uniform modeling methodology for benchmarking dnn accelerators.
\newblock In {\em 2019 IEEE/ACM International Conference on Computer-Aided
  Design (ICCAD)}, pages 1--7, 2019.

\bibitem{brevitas}
A.~Pappalardo.
\newblock Xilinx/brevitas.
\newblock \url{https://doi.org/10.5281/zenodo.3333552}.

\bibitem{NEURIPS2019_9015}
A.~Paszke, S.~Gross, F.~Massa, A.~Lerer, J.~Bradbury, G.~Chanan, T.~Killeen,
  Z.~Lin, N.~Gimelshein, L.~Antiga, A.~Desmaison, A.~Kopf, E.~Yang, Z.~DeVito,
  M.~Raison, A.~Tejani, S.~Chilamkurthy, B.~Steiner, L.~Fang, J.~Bai, and
  S.~Chintala.
\newblock Pytorch: An imperative style, high-performance deep learning library.
\newblock In H.~Wallach, H.~Larochelle, A.~Beygelzimer, F.~d\textquotesingle
  Alch\'{e}-Buc, E.~Fox, and R.~Garnett, editors, {\em Advances in Neural
  Information Processing Systems 32}, pages 8024--8035. Curran Associates,
  Inc., 2019.

\bibitem{rastegari2016xnor}
M.~Rastegari, V.~Ordonez, J.~Redmon, and A.~Farhadi.
\newblock Xnor-net: Imagenet classification using binary convolutional neural
  networks.
\newblock In {\em European conference on computer vision}, pages 525--542.
  Springer, 2016.

\bibitem{saad1990training}
D.~Saad and E.~Marom.
\newblock Training feed forward nets with binary weights via a modified chir
  algorithm.
\newblock {\em Complex Systems}, 4(5), 1990.

\bibitem{Sari2019HowDB}
E.~Sari, M.~Belbahri, and V.~P. Nia.
\newblock A study on binary neural networks initialization.
\newblock {\em CoRR}, abs/1909.09139, 2019.

\bibitem{simons2019review}
T.~Simons and D.-J. Lee.
\newblock A review of binarized neural networks.
\newblock {\em Electronics}, 8(6):661, 2019.

\bibitem{Simonyan14c}
K.~Simonyan and A.~Zisserman.
\newblock Very deep convolutional networks for large-scale image recognition.
\newblock In {\em International Conference on Learning Representations}, 2015.

\bibitem{singh2010visual}
G.~Singh and J.~Kosecka.
\newblock Visual loop closing using gist descriptors in manhattan world.
\newblock In {\em ICRA Omnidirectional Vision Workshop}, pages 44--48, 2010.

\bibitem{nordlands}
S.~Skrede.
\newblock Nordlandsbanen: minute by minute, season by season
  {https://nrkbeta.no/2013/01/15/nordlandsbanen-minute-by-minute-season-by-season/}.
\newblock Accessed: 2021-11-06.

\bibitem{sunderhauf2011brief}
N.~S{\"u}nderhauf and P.~Protzel.
\newblock Brief-gist-closing the loop by simple means.
\newblock In {\em 2011 IEEE/RSJ International Conference on Intelligent Robots
  and Systems}, pages 1234--1241. IEEE, 2011.

\bibitem{7353986}
N.~{Sünderhauf}, S.~{Shirazi}, F.~{Dayoub}, B.~{Upcroft}, and M.~{Milford}.
\newblock On the performance of convnet features for place recognition.
\newblock In {\em 2015 IEEE/RSJ International Conference on Intelligent Robots
  and Systems (IROS)}, pages 4297--4304, 2015.

\bibitem{tolias2016rmac}
G.~Tolias, R.~Sicre, and H.~J{\'e}gou.
\newblock {Particular Object Retrieval With Integral Max-Pooling of CNN
  Activations}.
\newblock In {\em {ICLR 2016}}, International Conference on Learning
  Representations, pages 1--12, San Juan, Puerto Rico, May 2016.

\bibitem{toneva2018an}
M.~Toneva, A.~Sordoni, R.~T. des Combes, A.~Trischler, Y.~Bengio, and G.~J.
  Gordon.
\newblock An empirical study of example forgetting during deep neural network
  learning.
\newblock In {\em International Conference on Learning Representations}, 2019.

\bibitem{umuroglu2017finn}
Y.~Umuroglu, N.~J. Fraser, G.~Gambardella, M.~Blott, P.~Leong, M.~Jahre, and
  K.~Vissers.
\newblock Finn: A framework for fast, scalable binarized neural network
  inference.
\newblock In {\em Proceedings of the 2017 ACM/SIGDA International Symposium on
  Field-Programmable Gate Arrays}, pages 65--74, 2017.

\bibitem{whaley2005minimizing}
R.~C. Whaley and A.~Petitet.
\newblock Minimizing development and maintenance costs in supporting
  persistently optimized blas.
\newblock {\em Software: Practice and Experience}, 35(2):101--121, 2005.

\bibitem{xia2019dnntune}
C.~Xia, J.~Zhao, H.~Cui, X.~Feng, and J.~Xue.
\newblock Dnntune: Automatic benchmarking dnn models for mobile-cloud
  computing.
\newblock {\em ACM Transactions on Architecture and Code Optimization (TACO)},
  16(4):1--26, 2019.

\bibitem{zaffar2020cohog}
M.~Zaffar, S.~Ehsan, M.~Milford, and K.~McDonald-Maier.
\newblock Cohog: A light-weight, compute-efficient, and training-free visual
  place recognition technique for changing environments.
\newblock {\em IEEE Robotics and Automation Letters}, 5(2):1835--1842, 2020.

\bibitem{memorable_maps}
M.~Zaffar, S.~Ehsan, M.~Milford, and K.~D. McDonald-Maier.
\newblock Memorable maps: A framework for re-defining places in visual place
  recognition.
\newblock {\em IEEE Transactions on Intelligent Transportation Systems}, pages
  1--15, 2020.

\bibitem{zaffar2020vpr}
M.~Zaffar, S.~Garg, M.~Milford, J.~Kooij, D.~Flynn, K.~McDonald-Maier, and
  S.~Ehsan.
\newblock Vpr-bench: An open-source visual place recognition evaluation
  framework with quantifiable viewpoint and appearance change.
\newblock {\em International Journal of Computer Vision}, pages 1--39, 2021.

\bibitem{zaffar2019state}
M.~Zaffar, A.~Khaliq, S.~Ehsan, M.~Milford, K.~Alexis, and K.~D.
  McDonald{-}Maier.
\newblock Are state-of-the-art visual place recognition techniques any good for
  aerial robotics?
\newblock {\em CoRR}, abs/1904.07967, 2019.

\bibitem{zaffar2019levelling}
M.~Zaffar, A.~Khaliq, S.~Ehsan, M.~Milford, and K.~McDonald-Maier.
\newblock Levelling the playing field: A comprehensive comparison of visual
  place recognition approaches under changing conditions.
\newblock In {\em IEEE ICRA Workshop on Dataset Generation and Benchmarking of
  SLAM Algorithms for Robotics and VR/AR}. IEEE, 2019.

\bibitem{zhou2017places}
B.~Zhou, A.~Lapedriza, A.~Khosla, A.~Oliva, and A.~Torralba.
\newblock Places: A 10 million image database for scene recognition.
\newblock {\em IEEE transactions on pattern analysis and machine intelligence},
  40(6):1452--1464, 2017.

\bibitem{zhou2016dorefa}
S.~Zhou, Y.~Wu, Z.~Ni, X.~Zhou, H.~Wen, and Y.~Zou.
\newblock Dorefa-net: Training low bitwidth convolutional neural networks with
  low bitwidth gradients.
\newblock {\em CoRR}, abs/1606.06160, 2016.

\bibitem{zhu2016trained}
C.~Zhu, S.~Han, H.~Mao, and W.~J. Dally.
\newblock Trained ternary quantization.
\newblock {\em CoRR}, abs/1612.01064, 2016.

\bibitem{zhuang2019structured}
B.~Zhuang, C.~Shen, M.~Tan, L.~Liu, and I.~Reid.
\newblock Structured binary neural networks for accurate image classification
  and semantic segmentation.
\newblock In {\em Proceedings of the IEEE/CVF Conference on Computer Vision and
  Pattern Recognition}, pages 413--422, 2019.

\end{thebibliography}

\end{document}